\documentclass[onefignum,onetabnum]{siamonline220329}

\usepackage{enumerate}
\usepackage{multirow}

\usepackage{makecell}
\usepackage{booktabs}
\usepackage{longtable}

\usepackage{bm}
\usepackage{amsmath}
\usepackage{amstext}
\usepackage{amssymb}
\usepackage{bbm}
\usepackage{amsbsy}
\usepackage{mathtools}
    
    \newcommand{\beq}{\begin{equation}}
    \newcommand{\eeq}{\end{equation}}
    \newcommand{\beqnn}{\begin{equation*}}
    \newcommand{\eeqnn}{\end{equation*}}
        
    \def\bp {\boldsymbol{p}}
    \def\bu {\boldsymbol{u}}

    \def\bw {\boldsymbol{w}}
    \def\bx {\boldsymbol{x}}
    
    \def\bz {\boldsymbol{z}}

    \def\R {\mathbb{R}}

    \def\bPhi {\boldsymbol{\Phi}}

    \DeclarePairedDelimiterX{\infdivx}[2]{(}{)}{%
        #1\;\delimsize\|\;#2%
    }
    \newcommand{\kldiv}{D_{\mathrm{KL}}\infdivx}

    \newcommand{\norm}[1]{\left\Vert{#1} \right\Vert}
    \newcommand{\normg}[1]{\left\Vert{#1} \right\Vert_{2,g}}

    \newcommand{\normop}[1]{\left\Vert{#1} \right\Vert_{\mathrm{op}}}

    \newcommand{\matr}[1]{\bm{#1}}
    
    \newcommand{\interior}[1]{\mathrm{int~}{#1}}
    
    \def\dom {\mathrm{dom~}}    % Domain
    \def\interior {\mathrm{int}}  % Interior
    \DeclareMathAlphabet{\mymathbb}{U}{BOONDOX-ds}{m}{n}
    
    \def\Rn {\R^n}
    \def\Rm {\R^m}
    
    \def\ical {\mathcal{I}}
    
    \def\evp {\mathbb{E}_{p}}
    \def\evpkone {\mathbb{E}_{p_{k+1}}}
    \def\evd {\mathbb{E}_{\hat{\mathcal{D}}}}
    \def\evprior {\mathbb{E}_{p_{\text{prior}}}}
    
    \DeclareMathOperator*{\argmin}{arg\,min}

    \newcommand{\normsq}[1]{\left\Vert{#1} \right\Vert_{2}^{2}}
    \newcommand{\normtwo}[1]{\left\Vert{#1} \right\Vert_{2}}
    \newcommand{\normone}[1]{\left\Vert{#1} \right\Vert_{1}}
    \newcommand{\norminf}[1]{\left\Vert{#1} \right\Vert_{\infty}}

    \title{Efficient first-order algorithms for large-scale, non-smooth maximum entropy models with application to wildfire science}

\author{Gabriel P. Langlois\thanks{Courant Institute of Mathematical Sciences, New York
University, New York City, NY 10012, USA (\email{Gabriel.provencher@courant.edu})} \and Jatan Buch\thanks{Department of Earth and Environmental Engineering, Columbia University, New York City, NY 10027, USA (\email{jb4625@columbia.edu})} \and J\'er\^ome Darbon\thanks{Division of Applied Mathematics, Brown University, Providence, RI 02912, USA (\email{jerome\_darbon@brown.edu})}}

\begin{document}
\maketitle

\begin{abstract}
Maximum entropy (Maxent) models are a class of statistical models that use the maximum entropy principle to estimate probability distributions from data. Due to the size of modern data sets, Maxent models need efficient optimization algorithms to scale well for big data applications. State-of-the-art algorithms for Maxent models, however, were not originally designed to handle big data sets; these algorithms either rely on technical devices that may yield unreliable numerical results, scale poorly, or require smoothness assumptions that many practical Maxent models lack. In this paper, we present novel optimization algorithms that overcome the shortcomings of state-of-the-art algorithms for training large-scale, non-smooth Maxent models. Our proposed first-order algorithms leverage the Kullback-Leibler divergence to train large-scale and non-smooth Maxent models efficiently. For Maxent models with discrete probability distribution of $n$ elements built from samples, each containing $m$ features, the stepsize parameters estimation and iterations in our algorithms scale on the order of $O(mn)$ operations and can be trivially parallelized. Moreover, the strong $\ell_{1}$ convexity of the Kullback--Leibler divergence allows for larger stepsize parameters, thereby speeding up the convergence rate of our algorithms. To illustrate the efficiency of our novel algorithms, we consider the problem of estimating probabilities of fire occurrences as a function of ecological features in the Western US MTBS-Interagency wildfire data set. Our numerical results show that our algorithms outperform the state of the arts by one order of magnitude and yield results that agree with physical models of wildfire occurrence and previous statistical analyses of wildfire drivers.
\end{abstract}

% Short title (50 chars): Efficient algorithms for non-smooth Maxent models



\section{Introduction}
Maximum entropy (Maxent) models are a class of density estimation methods that use the maximum entropy principle~\cite{jaynes1957information} to reproduce key statistics of datasets. Historically used in physics~\cite{jaynes1957information,jaynes1957informationII}, engineering~\cite{kapur1989maximum,gu2005detecting,berezinski2015entropy} and statistics~\cite{wainwright2008graphical} applications, Maxent models are now frequently used for large-scale machine learning problems in natural language processing~\cite{berger1996maximum,chen2000survey,della1997inducing,malouf2002comparison,Ratnaparkhi2017,tsujii2003evaluation}, social science~\cite{hayes2008maximum,kwon2004tmc,muttaqin2019maxent}, neuroscience~\cite{granot2013stimulus,tkavcik2013simplest,savin2017maximum}, ecological modeling~\cite{dudik2004maxent-performance,dudik2007maximum,elith2011statistical,kalinski2019building,merow2013practical,phillips2004maximum,phillips2006maximum,phillips2017opening,schnase2021toward,schnase2022automatic}, and wildfire science~\cite{Parisien2009, chen2021,yu_et_al_2021}.

Maxent models in machine learning must often estimate probability distributions from big datasets comprising hundreds of thousands to billions of samples or features or both~\cite{l2017machine}. Due to this enormous number, large-scale Maxent models need efficient and robust algorithms to perform well. State-of-the-art algorithms for Maxent models, however, were not originally designed to handle big datasets; these algorithms either rely on technical devices that may yield unreliable numerical results~\cite{fithian2013finite}, scale poorly in size~\cite{darroch1972generalized,della1997inducing}, or require smoothness assumptions that many Maxent models lack in practice~\cite{malouf2002comparison}. These limitations make it essentially impossible to scale Maxent models to big data applications without adequate and costly computational resources~\cite{demchenko2013addressing, thompson2021deep}. This constraint on computational resources, in particular, has been recently identified as a crucial challenge to overcome for using large-scale Maxent models in climate change and wildfire studies within a reasonable amount of run time~\cite{dyderski_2018, schnase2021toward,schnase2022automatic,chen2021}.

\subsection*{Contributions of this paper}
In this paper, we present novel optimization algorithms that overcome the shortcomings of state-of-the-art algorithms used for training large-scale, non-smooth Maxent models. Our proposed algorithms are first-order accelerated nonlinear primal-dual hybrid gradient (NPDHG) algorithms, whose theory was provided by two authors of this paper in~\cite{langlois2021accelerated}, based on the Kullback--Leibler divergence. Using the Kullback--Leibler divergence over the classical proximal operator makes it possible to train large-scale and non-smooth Maxent models much more efficiently than the state of the arts. In particular, we show that for a Maxent model with discrete probability distribution of $n$ elements built from samples each containing $m$ features, the stepsize parameters estimation and iterations in our algorithms all scale on the order of $O(mn)$ operations. This significantly improves on the known complexity bound of $O(\min(m^2n,mn^2))$ operations for computing the optimal stepsize parameters of classical first-order optimization methods, such as the linear PDHG or forward-backward splitting methods. We also show, as a consequence, that for a given tolerance level $\epsilon > 0$, our algorithms provably compute solutions using on the order of $O(mn/\sqrt{\epsilon})$ or $O(mn/\log(1/\epsilon))$ operations, the order depending on the smoothness of the Maxent model and which are optimal with respect to the Nesterov class of optimal first-order methods~\cite{nesterov2004lectures}. Moreover, the computational bottleneck consists of matrix-vector multiplications, which can be trivially parallelized, and so our proposed algorithms exhibit scalable parallelism. Finally, we show that the strong convexity of the Kullback--Leibler divergence with respect to the $\ell_{1}$ norm naturally allows for significantly larger stepsize parameters, thereby substantially speeding up the convergence rate of our algorithms.

To illustrate the efficiency of our novel algorithms on large-scale problems, we present an application to wildfire science. Specifically, we consider the problem of estimating probabilities of fire occurrences across the western US as a function of ecological features. To do so, we fit elastic net, non-overlapping group lasso and $\ell_{\infty}$ Maxent models to a large number of hyperparameters using our proposed algorithms and the state-of-the-art forward-backward splitting and coordinate descent STRUCTMAXENT2 algorithms~\cite{beck2009fast,mohri2018foundations}. Our numerical results show that our algorithms outperform the two latter algorithms by at least one order of magnitude, and yield results that are in good agreement with physical models of wildfire occurrence~\cite{Parisien_2019, Mezuman2020} as well as previous statistical analyses of wildfire drivers~\cite{Parisien2009, chen2021,yu_et_al_2021, Buch2023}.

\subsection*{Organization of this paper}
This paper is organized as follows. In \cref{sec:prelim}, we present our setup, we describe how Maxent models work, and we review three popular non-smooth Maxent models that are challenging to train with big data sets: the elastic net, group lasso and $\ell_{\infty}$ regularized Maxent models. In \cref{sec:challenges}, we explain why large-scale, non-smooth Maxent models are particularly challenging to train from big datasets, and we describe the state-of-the-art algorithms for training these Maxent models and their limitations. In \cref{sec:methodology}, we describe our approach for estimating Maxent models with NPDHG optimization methods, derive explicit algorithms for regularized Maxent models, and explain why our algorithms overcome the limitations of state-of-the-art algorithms. In \cref{sec:maxent_numerics}, we present an application of our algorithms to wildfire science, where we train a large number of Maxent models on the Western US MTBS-Interagency (WUMI) wildfire data set~\cite{Juang2022} to estimate probabilities of fire occurrences as a function of ecological features. Finally, we review our results and outline directions for future work in \cref{sec:conclusion}.
\section{Preliminaries}\label{sec:prelim}
\subsection{Setup}\label{subsec:setup}
Suppose we receive $l$ independent, identically distributed samples $\{v_{1},...,v_{l}\} \subset \ical$ from an unknown distribution $\mathcal{D}$. We assume throughout, without of loss of generality, that the input space $\ical \equiv \{1,\dots,n\}$. In addition, suppose we have a prior probability distribution $p_{\text{prior}}$ on $\ical$ that encapsulates some prior knowledge about the samples or unknown distribution. Finally, suppose we have access to a set of features from the samples via a bounded feature map $\bPhi\colon \ical \to \Rm$. Then how do we estimate the unknown distribution $\mathcal{D}$ from the prior distribution $p_{\text{prior}}$, the samples $\{v_{1},...,v_{l}\}$ and the feature map $\bPhi$?

The maximum entropy principle offers a systematic way to answer this question. It states that the distribution that best estimates the unknown distribution $\mathcal{D}$ is the one that remains as close as possible to the prior probability $p_{\text{prior}}$ while matching the features $\{\bPhi(v_{1}),\dots,\bPhi(v_{l})\}$ exactly or as closely as possible, in some suitable sense. Specifically, let $\Delta_{n}$ denote the $n$-dimensional probability simplex and suppose $p_{\text{prior}} \in \text{int}~{\Delta_{n}}$ (i.e., $p_{\text{prior}}(j) > 0$ for every $j \in \{1,\dots,n\}$). We measure closeness of a probability distribution $p \in \Delta_{n}$ to the prior probability $p_{\text{prior}}$ using the Kullback--Leibler divergence $D_{\text{KL}}\colon \Delta_{n} \times \text{int}~\Delta_{n} \to \R$:
\begin{equation}\label{eq:kl_div}
\kldiv{p}{p_{\text{prior}}} = 
\sum_{j=1}^{n} p(j)\log\left(p(j)/p_{\text{prior}}(j)\right).
\end{equation}
Next, let $\hat{\mathcal{D}}$ denote the empirical distribution induced by samples $\{v_{1},...,v_{l}\}$:
\begin{equation}\label{eq:empirical_dist}
    \hat{\mathcal{D}}(j) = \frac{1}{l}\left|\{1 \leqslant i \leqslant l \mid v_{i} = j \}\right|.
\end{equation}
In addition, let $\evd[\bPhi]$ denote the empirical average of the features induced by the samples and let $\evp[\bPhi]$ denote the model average induced by a probability distribution $p \in \Delta_{n}$:
\[
\evd[\bPhi] = \sum_{j=1}^{n} \hat{\mathcal{D}}(j)\bPhi(j) \quad \text{and} \quad \evp[\bPhi] = \sum_{j=1}^{n} p(j)\bPhi(j).
\]
We measure how the averages $\evd[\bPhi]$ and $\evp[\bPhi]$ match using a function $H^{*}\colon \Rm \to \R \cup \{+\infty\}$ called the potential function. Maxent models combine the two measures described above and seek to minimize the sum
\begin{equation}\label{eq:general_maxent}
    \inf_{p \in \Delta_{n}} f(p;t) = \inf_{p \in \Delta_{n}} \left\{\kldiv{p}{p_{\text{prior}}} + t H^{*}\left(\frac{\evd[\bPhi] - \evp[\bPhi]}{t}\right)\right\}
\end{equation}
where $t > 0$ is a hyperparameter selected using data-driven methods, e.g., cross-validation.

\textbf{Standing assumptions:} The potential function is proper, lower semicontinuous, convex, and bounded from below by zero with $H^{*}(\boldsymbol{0})$ = 0. Under the standing assumptions above on the potential function, which are fairly general, the probability distribution $\hat{\mathcal{D}}$ is a feasible point of~\cref{eq:general_maxent}. Moreover, the Maxent estimation problem~\cref{eq:general_maxent} admits a unique global solution (see \cite[Page 35, Proposition 1.2]{ekeland1999convex}; uniqueness follows from the strong convexity of the Kullback--Leibler divergence with respect to the $\ell_{1}$ norm~\cite{beck2003mirror,csiszar1967information,kemperman1969optimum,kullback1967lower,pinsker1964information}).

\subsection{Dual formulation and optimality conditions}
The generalized Maxent problem~\cref{eq:general_maxent} admits a dual problem corresponding to regularized maximum a posteriori estimation~\cite{altun2006unifying}. To derive the dual problem, we first write the second term on the right hand side of~\cref{eq:general_maxent} in terms of its convex conjugate,
\[
t H^{*}\left(\frac{\evd[\bPhi] - \evp[\bPhi]}{t}\right) = \sup_{\bw \in \Rm} \left\{\left\langle \bw,\evd[\bPhi] - \evp[\bPhi] \right\rangle - t H(\bw)\right\},
\]
where we abuse notation and write the convex conjugate of $H^{*}$ as $H$. This lets us express problem~\cref{eq:general_maxent} in saddle-point form:
\begin{equation}\label{eq:saddle_maxent}
    \inf_{p \in \Delta_{n}}\sup_{\bw \in \Rm}\left\{\left\langle \bw, \evd[\bPhi] - \evp[\bPhi]\right\rangle - tH(\bw) + \kldiv{p}{p_{\text{prior}}} \right\}.
\end{equation}
Due to our assumptions on the potential function $H^*$, we can swap the infimum and supremum~\cite[Statement (4.1) on page 61]{ekeland1999convex}. Using the convex conjugate formula
\[
\inf_{p \in \Delta_{n}}\left\{\kldiv{p}{p_{\text{prior}}} - \left\langle\bw,\evp[\bPhi]\right\rangle\right\} = - \log\left[\sum_{j=1}^{n} p_{\text{prior}}(j)e^{\left\langle \bw, \bPhi(j) \right\rangle}\right],
\]
we obtain the dual problem of~\cref{eq:general_maxent}:
\begin{equation}\label{eq:dual_maxent}
    \sup_{\bw \in \Rm} \left\{\left\langle \bw, \evd[\bPhi] \right\rangle - tH(\bw) - \log\left[\sum_{j=1}^{n} p_{\text{prior}}(j)e^{\left\langle \bw, \bPhi(j) \right\rangle}\right]\right\}.
\end{equation}

The dual problem~\cref{eq:dual_maxent} is a regularized maximum likelihood estimation problem over the family of Gibbs distributions~\cite{altun2006unifying,mohri2018foundations}. It has at least one global solution~\cite[Proposition (1.2) on page 35]{ekeland1999convex}. Moreover, if $(p^{s},\bw^{s})$ denotes a pair of solutions to the Maxent estimation problem~\cref{eq:general_maxent} and its dual~\cref{eq:dual_maxent}, then this pair satisfies the optimality conditions
\begin{equation}\label{eq:opt_cond}
\evd[\bPhi]-\mathbb{E}_{p^{s}}[\bPhi] \in t\partial H(\bw^{s}) \quad \text{and} \quad p^{s}(j) = \frac{p_{\text{prior}}(j)e^{\langle\bw^{s},\bPhi(j)\rangle}}{\sum_{j=1}^{n}p_{\text{prior}}(j)e^{\langle\bw^{s},\bPhi(j)\rangle}},
\end{equation}
for every $j\in \{1,\dots,n\}$ where $\partial H(\bw^{s})$ denotes the subdifferential of $H$ at $\bw^{s}$.

\subsection{Examples of non-smooth Maxent models}\label{subsec:examples}
Maxent models differ in the choice of the prior distribution $p_{\text{prior}}$, hyperparameter $t$ and the potential function $H^{*}$. In practice, the prior distribution is often chosen to be uniform and the hyperparameter is chosen via data-driven methods. The choice of potential function depends on the application. The classical Maxent model uses the indicator function
\[
\bu \mapsto H^{*}(\bu) = \begin{dcases}
0,\; &\text{if} \; \bu = \boldsymbol{0}, \\
+\infty\; &\text{otherwise}.
\end{dcases}
\] 
This forces the model average $\evp[\bPhi]$ to be equal to the empirical average $\evd[\bPhi]$. However, this is often too restrictive because the averages are expected to be close, and not equal, with high probability. Forcing the model and empirical averages to be equal can lead to severe over-fitting issues with big data sets, and more flexibility is therefore desired.

We review below three different non-smooth Maxent models that offer this flexibility: the elastic net, group lasso and $\ell_{\infty}$-regularized Maxent models. We chose these models because they are used extensively in statistics and machine learning but are also challenging to train on big data sets due to their non-smoothness~\cite{dudik2007maximum,phillips2017opening,dewaskar2023robustifying,schnase2021toward}. These models will also be used for our numerical experiments and results in \cref{sec:maxent_numerics}.

\subsubsection*{Elastic net regularized Maxent models}
These models use for potential function the relaxation of the convex conjugate of the $\ell_{1}$-norm:
\[
\bu \mapsto H^{*}(\bu) \equiv \left(\frac{1-\alpha}{2}\normsq{\cdot} + \alpha\normone{\cdot}\right)^{*}(\bu) = \frac{1}{2(1-\alpha)}\sum_{i=1}^{m}\max(0,|u_{i}|-\alpha)^2
\]
where $\alpha \in (0,1)$. This yields the elastic net regularized maximum entropy estimation problem
\begin{equation}\label{eq:maxent_EN}
    \min_{p \in \Delta_{n}} \left\{\kldiv{p}{p_{\text{prior}}} + \frac{1}{2t(1-\alpha)}\sum_{i=1}^{m}\max\left(0,\left|[\evd[\bPhi]]_{i}-[\evp[\bPhi]]_{i}\right|-t\alpha\right)^2\right\}.
\end{equation}
This type of regularization is frequently used for feature selection in machine learning~\cite{zou2005regularization,tay2023elastic,tsujii2003evaluation,dudik2007maximum}. The corresponding dual problem is
\begin{equation}\label{eq:maxent_EN_dual}
    \sup_{\bw \in \Rm} \left\{\left\langle \bw, \evd[\bPhi] \right\rangle - t\left(\frac{(1-\alpha)}{2}\normsq{\bw} + \alpha\normone{\bw}\right) -\log\left[\sum_{j=1}^{n} p_{\text{prior}}(j)e^{\left\langle \bw, \bPhi(j) \right\rangle}\right]\right\}.
\end{equation}
Thanks to the elastic net penalty $\frac{(1-\alpha)}{2}\normsq{\bw} + \alpha\normone{\bw}$, problem \cref{eq:maxent_EN_dual} is strongly concave and has a unique solution. 

An important aspect of the elastic net penalty is that it promotes sparsity; that is, it leads to a solution with many entries equal to zero~\cite{Foucart2013,el2020comparative,zanon2020sparse}, the number depending on the hyperparameter $t$ and parameter $\alpha$. Sparsity is useful when seeking accurate solutions based only on a few features or when performing feature selection.

\subsubsection*{Group lasso regularized Maxent models}
These models are specified starting from the dual problem. Consider a partition $G$ of the components of a vector $\bw \in \Rm$ into $g$ different and possibly overlapping groups $[\bw_{1},\dots,\bw_{g}]$ with $\bw_{g} \in \R^{m_{g}}$ and $\cup_{g=1}^{G} \bw_{g} = \bw$. Let $\bw \mapsto \norm{\bw_{g}}_{2,g}$ denote the $\ell_{2}$ norm over the respective components of the $g^{\text{th}}$ group. Then the dual version of the group lasso regularized maximum entropy estimation problem is
\begin{equation}\label{eq:maxent_group_dual}
    \sup_{\bw \in \Rm} \left\{\langle\bw,\evd[\bPhi]\rangle - t\sum_{g=1}^{G}\sqrt{m_{g}}\normg{\bw_{g}} - \log\left[\sum_{j=1}^{n} p_{\text{prior}}(j)e^{\left\langle \bw, \bPhi(j) \right\rangle}\right]\right\}.
\end{equation}
Thanks to the group lasso penalty $\sum_{g=1}^{G}\sqrt{m_{g}}\normg{\bw_{g}}$, problem \cref{eq:maxent_group} is strictly concave and so has at least one global solution. The corresponding primal problem follows from the convex conjugate formula
\[
\bu \mapsto H^{*}(\bu) \equiv \left(\sum_{g=1}^{G}\sqrt{m_{g}}\normg{\cdot}\right)^{*}(\bu) 
= \begin{dcases}
0, \; \text{if} \; \normg{\bu_{g}} \leqslant \sqrt{m_{g}} \;\text{for}\; g =1,\dots,G,\\
+\infty, \; \text{otherwise,}
\end{dcases}
\]
and it is therefore given by
\begin{equation}\label{eq:maxent_group}
    \min_{p \in \Delta_{n}} \kldiv{p}{p_{\text{prior}}} \quad \text{s.t.} \quad \normg{\evd[\bPhi_{g}] - \evp[\bPhi_{g}]} \leqslant t\sqrt{m_{g}} \quad \text{for} \quad g=1,\dots,G.
\end{equation}

Similarly to the elastic net penalty, the group lasso penalty promotes sparsity in the global solutions of~\cref{eq:maxent_group_dual}, but it differs in that it sets different blocks of components $[\bw_{1},\dots,\bw_{g}]$ to zero at the same time.

\subsubsection*{$\ell_{\infty}$-regularized Maxent models}
These models use for potential function the convex conjugate of the $\ell_{\infty}$ norm,
\[
\bu \mapsto H^{*}(\bu) \equiv \norm{\bu}_{\infty}^{*} = \left\{\hat{\bu} \in \Rm \mid \normone{\bu-\hat{\bu}} \leqslant 1\right\},
\]
which is the characteristic set of the $\ell_{1}$ ball with unit radius. This yields the $\ell_{\infty}$-regularized maximum entropy estimation problem
\begin{equation}\label{eq:maxent_linf}
    \min_{p \in \Delta_{n}} \left\{\kldiv{p}{p_{\text{prior}}} + t\norminf{\frac{\evd{[\bPhi]} - \evp{[\bPhi]}}{t}}^{*}\right\}.
\end{equation}
This type of regularization is used for certain machine learning problems, including matrix factor decomposition, and in robust statistics applications (see, e.g.,~\cite{lee2010practical,dewaskar2023robustifying}). The corresponding dual problem is
\begin{equation}\label{eq:maxent_linf_dual}
    \sup_{\bw \in \Rm} \left\{\langle\bw,\evd[\bPhi]\rangle - t\norminf{\bw} - \log\left[\sum_{j=1}^{n} p_{\text{prior}}(j)e^{\left\langle \bw, \bPhi(j) \right\rangle}\right]\right\}
\end{equation}
Thanks to the $\ell_{\infty}$ norm, problem \cref{eq:maxent_linf_dual} is strictly concave and has at least one global solution.
\section{Related work}\label{sec:challenges}
\subsection{Large-scale sparse Maxent models: Computational challenges}\label{subsec:challenges}
Estimating a probability distribution from the Maxent model~\cref{eq:general_maxent} can be computationally prohibitive for big data sets. To illustrate this point, suppose the Maxent model~\cref{eq:general_maxent} with hyperparameter $t\geqslant 0$ has a global solution $p^{s}(t)$. Let $p^{\epsilon}(t) \in \Delta_{n}$ with $\epsilon > 0$ denote an $\epsilon$-approximate solution to the global solution $p^{s}(t)$, that is, the objective function $f$ in~\cref{eq:general_maxent} satisfies
\[
f(p^{\epsilon}(t);t) - f(p^{s}(t);t) < \epsilon.
\]
If the potential function $H^{*}$ is smooth (equivalently, its conjugate $H$ is strongly convex), the best achievable rate of convergence for computing $p^{\epsilon}(t)$ with $t>0$ in the Nesterov class of optimal first-order methods is linear $O(\log(1/\epsilon))$ in the number of operations~\cite{nesterov2004lectures}. If $H$ is not smooth, then the optimal convergence rate is sublinear $O(1/\sqrt{\epsilon})$ in the number of operations.

These rates, while optimal, require carefully fine-tuned stepsize parameters. In classical first-order optimization algorithms, these stepsize parameters are fine-tuned using a precise estimate of the largest singular value of the linear operator $\matr{A} \colon \Delta_{n} \to \Rm$ defined by
\begin{equation}\label{eq:matrix_features}
\matr{A}p = \sum_{j=1}^{n}p(j)\bPhi(j) = \evp[\bPhi].
\end{equation}
Unfortunately, computing the largest singular value of the linear operator $\matr{A}$ in~\cref{eq:matrix_features} accurately is often computationally expensive for large matrices due to its prohibitive cubic cost of $O(\min{(m^2n,mn^2)})$ operations~\cite{hastie2009elements}. In this situation, line search methods and other heuristics can sometime be employed to bypass this issue, but they typically slow down the convergence speed. Another approach is to compute a crude estimate of the largest singular value of the matrix $\matr{A}$, but doing so significantly reduces convergence speed as well. In fact, even if the largest singular value can be computed quickly, the resulting stepsize parameters may be much smaller than what is permissible to maintain convergence, that is, the largest singular value itself may be a poor estimate for determining how large the stepsize parameters are allowed to be to maintain convergence. This point will discussed in more detail in \cref{subsec:methodology} when we describe our proposed NPDHG algorithms.

This issue makes solving the Maxent model~\cref{eq:general_maxent} difficult and laborious. Even worse, in some applications, the appropriate value of the hyperparameter $t$ in~\cref{eq:general_maxent} is difficult to guess and must be selected by repeatedly solving~\cref{eq:general_maxent} from a large pool of values of hyperparameters, a process that can become particularly time-consuming and resource intensive for big data sets. This issue has driven much research in developing robust and efficient algorithms to minimize computational costs and maximize model performance.

\subsection{State-of-the-art methods for large-scale, non-smooth Maxent models}\label{sec:sota_methods}
State-of-the-art methods for computing solutions to large-scale, non-smooth Maxent models are based on coordinate descent algorithms~\cite{friedman2007pathwise,friedman2010regularization,hastie2021glmnet,cortes2015maxent} and first-order optimization algorithms such as forward-backward splitting~\cite{beck2009fast,chambolle2016introduction,daubechies2004iterative}.  We'll discuss these methods below, but note that other types of methods have been developed for Maxent models; see, e.g.,~\cite{dudik2006maximum,malouf2002comparison,mann2009efficient,mohri2018foundations} for surveys and comparisons of different algorithms. In particular, we will not consider second-order based methods suitable for smooth Maxent models such as limited-memory BFGS algorithms~\cite{malouf2002comparison,andrew2007scalable} since this work focuses on non-smooth Maxent models.

\noindent
\newline
\textbf{Coordinate descent methods.} The state-of-the-art is a coordinate descent algorithm based on a technical device called infinitely weighted logistic regression (IWLR)~\cite{fithian2013finite,phillips2017opening}. The IWLR method approximates a Maxent model as a logistic regression model and then fits the approximate logistic regression model using an existing, efficient optimization algorithm for logistic regression. This indirect approach was proposed because efficient and scalable coordinate descent algorithms were already available for fitting logistic regression models, which in~\cite{fithian2013finite} was an earlier version of the GLMNET software package~\cite{hastie2021glmnet}. The IWLR method remains the state-of-the-art for this reason. It is implemented, for example, in the Maxent package available in the R programming language~\cite{hastie2021glmnet,phillips2017opening}.

The IWLR method, however, is an approximate technical device that is not guaranteed to work, a fact acknowledged by the authors who proposed the method~\cite{fithian2013finite}, and therefore may not produce reliable numerical results. Coordinate descent algorithms themselves are generally non-parallelizable and often lack robustness and good convergence properties. For example, the aforementioned GLMNET software package approximates the logarithm term in logistic regression models with a quadratic term to fit the models efficiently. Without costly step-size optimization, which GLMNET avoids to improve performance, the GLMNET implementation may not converge~\cite{friedman2010regularization,lee2006efficient}. Case in point, \cite{yuan2010comparison} provides two numerical experiments in which GLMNET does not converge. 

Other coordinate descent algorithms have been developed to compute solutions to non-smooth Maxent models directly (see, e.g.,~\cite{yu2011dual,cortes2015maxent}) but they are also generally non-parallelizable and often lack robustness and good convergence properties. Finally, another issue is that many coordinate descent algorithms, including GLMNET, were designed for sparse Maxent models (e.g., the elastic net and group lasso regularized Maxent models described in \cref{subsec:examples}) and those algorithms depend on the sparsity of the model to converge quickly~\cite{zou2005regularization}. It would be desirable to have fast optimization methods for when the feature mapping~\cref{eq:matrix_features} is dense, as this often occurs in practice.

\noindent
\newline
\textbf{First-order methods.} First-order optimization algorithms such as the forward-backward splitting algorithm are popular because they are robust and can provably compute $\epsilon$-approximate solutions of~\cref{eq:general_maxent} with an optimal rate of convergence. However, as discussed in detail at the end of the previous subsection, achieving this rate of convergence requires fine-tuning the stepsize parameters using an accurate estimate of the largest singular value of the feature mapping~\cref{eq:matrix_features}, and this estimate is typically computationally expensive for large matrices. This problem makes classical first-order optimization algorithms inefficient and impractical in estimating probability densities from large-scale Maxent models.

\noindent
\newline
\textbf{Summary.} The current state-of-the-art algorithms for estimating probability densities from Maxent models either produce unreliable numerical results, lack scalable parallelism or scale poorly in size. These shortcomings in terms of robustness and efficiency make it challenging to use non-smooth, large-scale Maxent models in applications without access to adequate and costly computational resources. The next section presents novel, efficient and robust accelerated NPDHG optimization methods that address these shortcomings.
\section{Main results}\label{sec:methodology}
We describe here our approach for computing solutions to the generalized Maxent problem~\cref{eq:general_maxent} using accelerated NPDHG optimization methods~\cite{langlois2021accelerated}. In addition to the standing assumptions on the potential function $H^{*}$, we assume we can compute efficiently the proximal operator of the convex conjugate of the potential function
\[
\argmin_{\bw \in \Rm} \left\{\lambda H(\bw) + \frac{1}{2}\normsq{\bw-\hat{\bw}}\right\},
\]
for every $\lambda > 0$ and $\hat{\bw} \in \Rm$. These assumptions are satisfied for most potential functions used in practice, including for the potential functions described in \cref{subsec:examples}.

\subsection{Methodology}\label{subsec:methodology}
We start with the saddle-point formulation~\cref{eq:saddle_maxent} of the generalized Maxent estimation problem~\cref{eq:general_maxent}. Based on the work two of the authors provided in~\cite{langlois2021accelerated}, we propose to split the infimum and supremum in the saddle-point problem~\cref{eq:saddle_maxent} using an iterative NPDHG scheme that alternates between a nonlinear proximal ascent step based on the Kullback--Leibler divergence and a linear proximal descent step.

More precisely, let $\tau_{0} > 0$ and $\sigma_{0} > 0$ be two stepsize parameters satisfying the inequality
\begin{equation}\label{eq:ineq_parameters}
\tau_{0}\sigma_{0} \left(\max_{j \in \{1,\dots,n\}}\normsq{\bPhi(j)}\right)\leqslant 1,
\end{equation}
let $\theta_{0} = 0$, let $\bw_{-1} = \bw_{0} \in \interior{~\dom H}$, let $\bz_{0} \in \Rm$ and define the initial probability distribution $p_{0} \in \Delta_{n}$ through the constant $\bz_{0}$ via
\begin{equation}\label{eq:p0}
p_{0}(j) = \frac{p_{\text{prior}}(j)e^{\langle\bz_{0},\bPhi(j)\rangle}}{\sum_{j=1}^{n}p_{\text{prior}}(j)e^{\langle\bz_{0},\bPhi(j)\rangle}}
\end{equation}
for each $j \in \{1,\dots,n\}$. Our proposed NPDHG iterative scheme computes the iterates
\begin{equation}\label{eq:NPDHG_g-scheme_nonsmooth}
\begin{alignedat}{1}
p_{k+1} &= \argmin_{p \in \Delta_{n}} \left\{\tau_{k}\kldiv{p}{p_{\text{prior}}} -\tau_{k}\left\langle \bw_{k} + \theta_{k}(\bw_{k}-\bw_{k-1}),\evp[\bPhi]\right\rangle + \kldiv{p}{p_{k}} \right\}\\
\hat{\bw}_{k+1} &= \bw_{k} + \sigma_{k}(\evd[\bPhi] - \evpkone[\bPhi]),\\
\bw_{k+1} &= \argmin_{\bw \in \Rm} \left\{t\sigma_{k}H(\bw) + \frac{1}{2}\normsq{\bw-\hat{\bw}_{k+1}}\right\},\\
\theta_{k+1} &= 1/\sqrt{1+\tau_{k}},\quad \tau_{k+1} = \theta_{k+1}\tau_{k}\quad \text{and} \quad \sigma_{k+1} = \sigma_{k}/\theta_{k+1}.
\end{alignedat}
\end{equation}
According to~\cite[Proposition 4.1]{langlois2021accelerated}, the sequence of iterates $p_{k}$ converges strongly to the unique solution of the generalized Maxent problem~\cref{eq:general_maxent}. Moreover, for any $t\geqslant 0$ and a given tolerance level $\epsilon > 0$, the scheme~\cref{eq:NPDHG_g-scheme_nonsmooth} provably computes an $\epsilon$-approximate solution $p^{\epsilon}(t)$ of the generalized Maxent model~\cref{eq:general_maxent} in $O(1/\sqrt{\epsilon})$ time. This rate of convergence is, without further smoothness assumptions on the potential function $H^{*}$, the best achievable rate of convergence with respect to the Nesterov class of optimal first-order methods~\cite{nesterov2004lectures}.

The key element in this scheme is the choice of the Kullback--Leibler divergence as a nonlinear proximal step in the first line of~\cref{eq:NPDHG_g-scheme_nonsmooth}. We use it for two reasons: First, because the Kullback--Leibler divergence already appears in the saddle-point problem~\cref{eq:saddle_maxent}. This allows us to compute $p_{k+1}$ explicitly. Indeed, thanks to the choice of initial probability distribution~\cref{eq:p0}, we have
\begin{equation}\label{eq:update_pkplusone}
p_{k+1}(j) = \frac{p_{\text{prior}}(j)e^{\langle\bz_{k+1},\bPhi(j)\rangle}}{\sum_{j=1}^{n}p_{\text{prior}}(j)e^{\langle\bz_{k+1},\bPhi(j)\rangle}} \; \text{with} \; \bz_{k+1} = \frac{1}{1+\tau_{k}}\left(\bz_{k} + \tau_{k}(\bw_{k} + \theta_{k}(\bw_{k}-\bw_{k-1}))\right)
\end{equation}
for each $j \in \{1,\dots,n\}$. See \cref{supp:update-p} for the derivation of~\cref{eq:update_pkplusone}.

Second, because the Kullback--Leibler divergence is $1$-strongly convex with respect to the $\ell_{1}$ norm, that is,
\[
\kldiv{p}{p_{k}} \geqslant \frac{1}{2}\norm{p-p_{k}}_{1}^{2}.
\]
This fact follows from a fundamental result in information theory known as Pinsker's inequality~\cite{beck2003mirror,csiszar1967information,kemperman1969optimum,kullback1967lower,pinsker1964information}. In particular, this means that the scheme~\cref{eq:NPDHG_g-scheme_nonsmooth} alternate between solving a strongly convex problem over the space $(\Rn,\normone{\cdot})$ and a concave problem over the space $(\Rm,\normtwo{\cdot})$. The choice of these spaces is significant, for the induced operator norm $\normop{\cdot}$ of the linear operator $\matr{A}$ defined in~\cref{eq:matrix_features} becomes
\[
\normop{\matr{A}} = \sup_{\normone{p} = 1}\normtwo{\matr{A}p} = \max_{j \in \{1,\dots,n\}}\normtwo{\bPhi(j)}.
\]
This operator norm offers two crucial advantages: First, it can be computed in optimal $\Theta(mn)$ time, or better if the features $\{\bPhi(j)\}_{j=1}^{n}$ have structure, e.g., if they are sparse. This means that the stepsize parameters $\tau_{0}$ and $\sigma_{0}$ of the NPDHG scheme can be computed in~\cref{eq:ineq_parameters}, with equality, in optimal $\Theta(mn)$ time. This is important because in classical first-order optimization methods, we typically require a precise estimate estimate of the largest singular value of the feature mapping~\cref{eq:matrix_features}, namely the number
\[
\norm{\matr{A}}_{2} = \sup_{\normtwo{\bx} = 1}\norm{\matr{A}p}_{2},
\]
to fine-tune the stepsize parameters to gain an optimal convergence rate. However, as discussed in detail in \cref{subsec:challenges}, this estimate is computationally expensive for large matrices due to its prohibitive cubic cost of $O(\min{(m^2 n,mn^2)})$ operations. In contrast, our NPDHG scheme~\cref{eq:NPDHG_g-scheme_nonsmooth} does not suffer from this computational bottleneck and therefore scales well. Second, the operator norm $\normop{\matr{A}}$ can be significantly smaller than the estimate $\norm{\matr{A}}_{2}$, hence allowing for bigger stepsize parameters to further speed up convergence\footnote{An easy calculation yields $\max_{\normtwo{\bp}=1}\normtwo{\matr{A}\bp} \geqslant \max_{j \in \{1,\dots,n\}} \norm{\bPhi(j)}_{2}$.}.

\noindent
\newline
\textbf{Summary.} To solve the generalized Maxent estimation problem~\cref{eq:general_maxent} and its dual problem~\cref{eq:dual_maxent}, let $\tau_{0} > 0$ and $\sigma_{0} > 0$ be two stepsize parameters satisfying inequality~\cref{eq:ineq_parameters}, let $\theta_{0} = 0$, let $\bw_{-1} = \bw_{0} \in \interior{~\dom H}$, let $\bz_{0} \in \Rm$ and define an initial probability distribution $p_{0}$ through $\bz_{0}$ via~\cref{eq:p0}. Then compute the iterates
\begin{equation}\label{eq:NPDHG_g-scheme_nonsmooth_explicit}
\begin{alignedat}{1}
\bz_{k+1} &= \left(\bz_{k} + \tau_{k}(\bw_{k} + \theta_{k}(\bw_{k}-\bw_{k-1}))\right)/(1+\tau_{k}),\\
p_{k+1}(j) &= \frac{p_{\text{prior}}(j)e^{\langle\bz_{k+1},\bPhi(j)\rangle}}{\sum_{j=1}^{n}p_{\text{prior}}(j)e^{\langle\bz_{k+1},\bPhi(j)\rangle}} \quad \text{for} \; j\in\{1,\dots,n\},\\
\hat{\bw}_{k+1} &= \bw_{k} + \sigma_{k}(\evd[\bPhi] - \evpkone[\bPhi]),\\
\bw_{k+1} &= \argmin_{\bw \in \Rm} \left\{t\sigma_{k}H(\bw) + \frac{1}{2}\normsq{\bw-\hat{\bw}_{k+1}}\right\},\\
\theta_{k+1} &= 1/\sqrt{1+\tau_{k}},\quad \tau_{k+1} = \theta_{k+1}\tau_{k}\quad \text{and} \quad \sigma_{k+1} = \sigma_{k}/\theta_{k+1},
\end{alignedat}
\end{equation}
until convergence is achieved. As all parameters and updates can be computed in $O(mn)$ time, for any $t\geqslant 0$ and a given tolerance level $\epsilon > 0$ the overall complexity for computing an $\epsilon$-approximate solution $p^{\epsilon}(t)$ is $O(mn/\sqrt{\epsilon})$.

\subsection{Algorithm for smooth potential functions}
The iterative scheme~\cref{eq:NPDHG_g-scheme_nonsmooth_explicit} does not require the potential function $H^{*}$ to be smooth. If, however, the potential function $H^{*}$ is $\gamma_{H^{*}}$-smooth (equivalently, $H$ is $\frac{1}{\gamma_{H^{*}}}$-strongly convex) for some $\gamma_{H^{*}} > 0$, then we can modify the NPDHG iterative scheme~\cref{eq:NPDHG_g-scheme_nonsmooth_explicit} to achieve a linear rate of convergence. More precisely, let $t>0$ and introduce the stepsize parameters
\begin{equation}\label{eq:parameters_smooth}
    \theta = 1  - \frac{t}{2\gamma_{H^{*}}\normop{\matr{A}}^{2}}\left(\sqrt{1+\frac{4\gamma_{H^{*}}\normop{\matr{A}}^{2}}{t}}-1\right), \quad \tau = \frac{1-\theta}{\theta} \quad \mathrm{and} \quad \sigma = \frac{\gamma_{H^{*}}\tau}{t}.
\end{equation}
Let $\bw_{-1} = \bw_{0} \in \interior{~\dom H}$, let $\bz_{0} \in \Rm$ and define $p_{0}$ through $\bz_{0}$ via~\cref{eq:p0}. Then the explicit NPDHG iterative scheme is
\begin{equation}\label{eq:NPDHG_g-scheme_smooth_explicit}
\begin{alignedat}{1}
\bz_{k+1} &= \left(\bz_{k} + \tau(\bw_{k} + \theta(\bw_{k}-\bw_{k-1}))\right)/(1+\tau),\\
p_{k+1}(j) &= \frac{p_{\text{prior}}(j)e^{\langle\bz_{k+1},\bPhi(j)\rangle}}{\sum_{j=1}^{n}p_{\text{prior}}(j)e^{\langle\bz_{k+1},\bPhi(j)\rangle}} \quad \text{for} \; j\in\{1,\dots,n\},\\
\hat{\bw}_{k+1} &= \bw_{k} + \sigma(\evd[\bPhi] - \evpkone[\bPhi]),\\
\bw_{k+1} &= \argmin_{\bw \in \Rm} \left\{t\sigma H(\bw) + \frac{1}{2}\normsq{\bw-\hat{\bw}_{k+1}}\right\}.\\
\end{alignedat}
\end{equation}
According to~\cite[Proposition 4.3]{langlois2021accelerated}, the sequences of iterates $p_{k}$ and $\bw_{k}$ converge strongly to the unique solution of the generalized Maxent estimation problem~\cref{eq:general_maxent} and its dual problem~\cref{eq:dual_maxent}. Moreover, for any $t > 0$ and a given tolerance level $\epsilon > 0$, this scheme provably computes an $\epsilon$-approximate solution $p^{\epsilon}(t)$ of the generalized Maxent estimation problem~\cref{eq:general_maxent} in $O(\log(1/\epsilon))$. This rate of convergence is the best achievable rate of convergence with respect to the Nesterov class of optimal first-order methods~\cite{nesterov2004lectures}. As all parameters and updates can be computed in $O(mn)$ time, the overall complexity for computing an $\epsilon$-approximate solution $p^{\epsilon}(t)$ is $O(mn\log(1/\epsilon))$.
\section{Application to wildfire science}\label{sec:maxent_numerics}
To illustrate the efficiency of our novel algorithms on large-scale problems, we present here an application to wildfire science. The problem at hand is to combine fire occurrence data with ecological data in a fixed geographical region to estimate the probability of fire occurrences as a function of ecological features. Maxent models achieve this goal by translating fire occurrence and ecological data into probabilities of fire occurrences and ecological features. This approach closely mirrors how Maxent models are used for modeling species geographic distributions~\cite{dudik2004maxent-performance,dudik2007maximum,kalinski2019building,phillips2004maximum,phillips2006maximum,phillips2017opening,schnase2021toward,schnase2022automatic}. Another related goal is to identify what ecological features correlate most with fire occurrences. This can be achieved using a sparse Maxent model, e.g., an elastic net or group lasso regularized Maxent model, to identify ecological features correlating significantly with fire occurrences.

For this application, we use the Western US MTBS-Interagency (WUMI) wildfire data set~\cite{Juang2022}, which we describe in \cref{subsec:wumi_climate_data_set} below. We formulate the problem of combining the fire occurrence and ecological data from the WUMI wildfire data set into a Maxent estimation problem in \cref{subsec:data_to_maxent}. Using this data, we then fit the elastic net, (non-overlapping) group lasso and $\ell_{\infty}$ Maxent models for a large number of hyperparameters weighting the regularization. We detail this fitting procedure and the explicit NPDHG algorithms for these Maxent models in \cref{subsec:fitting_setup}. Following this, we compare in \cref{subsec:numerics_comparison} the running times required to fit the aforementioned Maxent models using our NPDHG algorithms with the forward-backward splitting algorithm~\cite{beck2009fast,chambolle2016introduction} and the STRUCTMAXENT2 coordinate descent algorithm of~\cite{mohri2018foundations}. Finally, in \cref{subsec:numerics_analysis}, we interpret the results obtained from fitting the aforementioned Maxent models to the WUMI wildfire data set.

\subsection{WUMI wildfire data set}\label{subsec:wumi_climate_data_set}
\begin{figure}[!t]
\centering
\includegraphics[width= 0.95\textwidth]{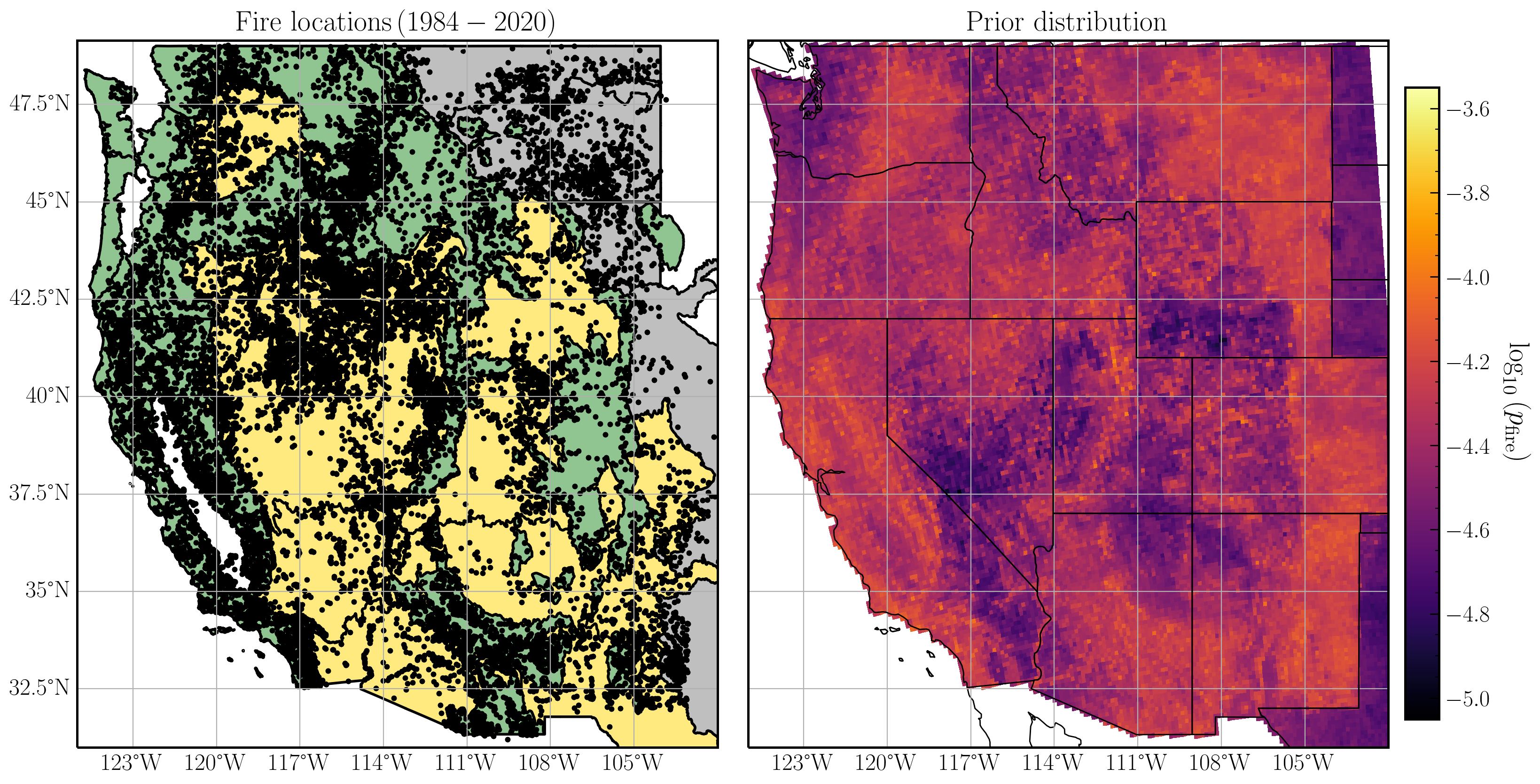}
\caption{Wildfire activity in the western United States from 1984 to 2020. (Left) Fire locations of all fires (black dots) in the Western US MTBS-Interagency (WUMI) data set; also shown are three ecological divisions characterized by their primary vegetation type – forests (green), deserts (yellow), and plains (gray). (Right) Prior distribution indicating mean fire probability across all calendar months.
}
\label{fig:introplot}
\end{figure}

The Western US MTBS-Interagency wildfire data set~\cite{Juang2022} consists of all fires ($\geqslant 1 \, {\rm km}^2$) from multiple state and federal agencies, supplemented by satellite observations of large fires ($\geqslant 4 \, {\rm km}^2$) from the Monitoring Trends in Burn Severity (MTBS) program, in the continental United States west of $103^\circ$ W longitude. For this application, we extracted all wildfires from the WUMI data set that occurred between 1984-2020 inclusive (accessed May 18, 2023). The locations of all fires used are shown in \cref{fig:introplot}.

Next, following the procedure outlined in~\cite{Buch2023}, we overlay the fire locations on a $12 \, {\rm km} \times 12 \, {\rm km}$ grid to construct a data frame of prevailing climate, vegetation, topographic, and human-related features for each fire. Altogether we include a total of 35 potential fire-related features from various sources; we provide a summary of all features used in our analysis along with their sources in \cref{supp:wumi_features}. For each grid cell, we also provide an index identifying its Environmental Protection Agency (EPA) Level III ecoregion. Defined on the basis of roughly homogeneous climate and vegetation patterns, ecoregions are commonly used in the wildfire science literature to identify climate-fire relationships at a coarse spatial scale~\cite{littell_2009, Parisien2009}. The full WUMI wildfire data set for all months from 1984-2020 is publicly available~\cite{buch_2022} as part of the Stochastic Machine Learning for wildfire activity in the western US (SMLFire1.0) package.

\subsection{Data preprocessing}\label{subsec:data_to_maxent}
We construct a weakly informative prior~\cite{mcelreath2018, gelman2013bayesian} for incorporating existing knowledge of wildfire conducive environmental conditions in our Maxent model. First, we prepare a training data set of 10000 fire occurrences and absences chosen randomly across all months between 1984-2020 correlated with the values of 35 fire-related features described in the previous section. The features are averaged over each calendar month (i.e., January, February, ...) between 1984 and 2020. We then apply min-max scaling to each feature, ensuring that all features lie in the same range. Second, we construct two Random Forest (RF) models, one each for fires in dry (May -- September) and wet (October -- April) seasons respectively. We withhold 20$\%$ of the training data to tune the model hyperparameters, such that the optimal hyperparameters ensure a trade-off between model precision and recall. Next, we predict the fire probability for all grid cells with fire-related features using either the trained wet or dry season RF model depending on the calendar month. Since the Maxent algorithm assumes a presence-only framework, that is, the absence of fires in a grid cell does not imply a non-zero fire probability, we impute grid cells without a fire with probability, $p_{\rm nfire} = {\rm min}(p_{\rm fire})/10$. Last, we normalized the prior probability distribution to ensure that the fire probability across all grid cells sums up to one. For convenience, we represent the prior distribution as the normalized mean of 12 monthly fire probabilities predicted by the RF models in \cref{fig:introplot}. 

To construct the empirical distribution $\hat{\mathcal{D}}$, we divide the study region into its 18 different EPA level III ecoregions and weigh the relative frequencies of fire among the ecoregions using the following strategy. For each ecoregion $r \in \{1,\dots,18\}$, let $n_{\text{r,total}}$ denote the total number of grid cells in ecoregion $r$ and let $n_{\text{r,fire}}$ denote the total number grid cell in ecoregion $r$ where at least one fire was recorded. In addition, let $Z = \sum_{r=1}^{18} \frac{n_{\text{r,fire}}}{n_{\text{r,total}}}$ denote the sum of the proportions of grid cells where at least one fire was recorded among all ecoregions. Then we compute the empirical probability at cell $j \in \{1,\dots,n\}$, $\hat{\mathcal{D}}(j)$, as follows:
\begin{equation}\label{eq:emp_prob_calc}
    \hat{{\mathcal{D}}}(j) = \frac{1}{Z} \begin{cases}
        0 &\; \text{if no fire was recorded in grid cell $j$}, \\
        1/n_{\text{r,total}} &\; \parbox[t]{.6\textwidth}{if at least one fire was recorded in grid $j$ and the grid cell $j$ belongs to ecoregion $r$}
    \end{cases}
\end{equation}
This construction gives the empirical distribution more weight to ecoregions where fires are more frequent and widespread.

\subsection{Fitting procedure and algorithmic setup}\label{subsec:fitting_setup}
For the analysis, we fit six different Maxent models to the wildfire data: four elastic net Maxent models with parameters $\alpha = \{0.95,0.4,0.15,0.05\}$, a non-overlapping group lasso Maxent model, and an $\ell_{\infty}$-regularized Maxent model. For each model, we fit a regularization path of 141 hyperparameters as follows:
\[
t^{(l)} = 
\begin{cases}
(1-l/100)t^{(0)} &\; \text{from}\; l = 0\,\text{to}\,50 \\
(0.5-(l-50)/200)t^{(0)} &\; \text{from}\; l = 51\,\text{to}\, 140,
\end{cases}
\]
where $t^{(0)}$ depends on the Maxent model and corresponds to the smallest hyperparameter for which the primal and dual solutions are equal to the prior distribution and zero. More precisely, for each model we set $(p^{(0)},\bw^{(0)}) = (p_{\text{prior}},\boldsymbol{0})$ and compute the sequence of solutions $\{(p^{(l)},\bw^{(l)})\}_{l=1}^{140}$ of the corresponding Maxent primal and dual problems \cref{eq:maxent_EN} and \cref{eq:maxent_EN_dual} with hyperparameter $t = t^{(l)}$.

We chose these Maxent models to study the impact of the regularization on the primal and dual solutions as a function of the hyperparameters. In particular, for the elastic net and non-overlapping group lasso Maxent models, we are interested in identifying the set of features that are selected or discarded as a function of the sequence of hyperparameters $\{t^{(l)}\}_{l=0}^{140}$. In the parts below, we describe the value of $t^{(0)}$ and the NPDHG algorithm used for each model.

\subsubsection*{Elastic net Maxent models}
For these Maxent models, the smallest hyperparameter for which the solutions to the primal and dual problems \cref{eq:maxent_EN} and \cref{eq:maxent_EN_dual} are equal to the prior distribution and zero is
\[
t^{(0)} = \norminf{\evd[\bPhi]-\evprior[\bPhi]}/\alpha.
\]
This follows from the optimality condition
\begin{equation}\label{eq:opt_cond_elastic}
\evd[\bPhi]-\mathbb{E}_{p^{s}}[\bPhi] - t(1-\alpha)\bw^{s} \in t\alpha\partial\normone{\cdot}\bw^{s}
\end{equation}
and computing the smallest parameter $t$ for which \cref{eq:opt_cond_elastic} is satisfied at $(p^{s},\bw^{s}) = (p_{\text{prior}},\boldsymbol{0})$.

For the elastic net Maxent model with $\alpha = 0.95$, we use the NPDHG algorithm \cref{eq:NPDHG_g-scheme_nonsmooth_explicit} (with sublinear convergence rate), while for the models with $\alpha = \{0.4,0.15,0.05\}$, we use the NPDHG algorithm \cref{eq:NPDHG_g-scheme_smooth_explicit} (with linear convergence rate). Starting from $l = 1$, we compute the pair of solutions $(p^{(l)},\bw^{(l)})$ to the primal and dual problems \cref{eq:maxent_EN} and \cref{eq:maxent_EN_dual} at hyperparameter $t = t^{(l)}$ using the previously computed pair of solutions $(p^{(l-1)},\bw^{(l-1)})$ by setting the initial vectors $\bz_{0} = \bw_{0} = \bw^{(l-1)}$ in both NPDHG algorithms. For the stepsize parameters, we set $\theta_{0} = 0$, $\tau_{0} = 2$ and $\sigma_{0} = 1/2\normop{\matr{\norm{A}}}^{2}$ in \cref{eq:NPDHG_g-scheme_nonsmooth_explicit}, and we set $\theta$, $\tau$ and $\sigma$ according to the formulas in \cref{eq:parameters_smooth} with $\gamma_{H^{*}} = 1-\alpha$ in \cref{eq:NPDHG_g-scheme_smooth_explicit}. We compute the update $\bw_{k+1}$ in both NPDHG algorithms using the classical soft thresholding operator~\cite{daubechies2004iterative,figueiredo2001wavelet,lions1979splitting}. Specifically, for any $\lambda > 0$, $\hat{\bw} \in \Rm$ and $i \in \{1,\dots,m\}$ we have
\begin{equation*}
\left[\argmin_{\bw \in \Rm} \left\{\lambda\normone{\bw} + \frac{1}{2}\normsq{\bw-\hat{\bw}}\right\}\right]_{i} \equiv \left[\text{shrink}_{1}(\hat{\bw},\lambda)\right]_{i} =
\begin{dcases}
    [\hat{\bw}]_{i} - \lambda &\quad \mathrm{if}\, [\hat{\bw}]_{i} > \lambda, \\
    0 &\quad \mathrm{if}\, |[\hat{\bw}]_{i}| \leqslant \lambda, \\
    [\hat{\bw}]_{i} + \lambda &\quad \mathrm{if}\, [\hat{\bw}]_{i} < -\lambda,
\end{dcases}
\end{equation*}
and so, for every $\lambda > 0$, $\alpha \in [0,1]$, $\hat{\bw} \in \Rm$ and $i \in \{1,\dots,m\}$ we have
\begin{equation*}
\left[\argmin_{\bw \in \Rm} \left\{\lambda\left(\alpha\normone{\bw} + \frac{(1-\alpha)}{2}\normsq{\bw}\right) + \frac{1}{2}\normsq{\bw-\hat{\bw}}\right\}\right]_{i} = \frac{[\text{shrink}_{1}(\hat{\bw},\lambda\alpha)]_{i}}{1+\lambda(1-\alpha)}.
\end{equation*}
Finally, we let the NPDHG algorithms run for at least 40 iterations before checking for convergence. We stop the NPDHG algorithms when the optimality condition \cref{eq:opt_cond_elastic} is satisfied within some tolerance $10^{-5}$:
\[
\norminf{\evd[\bPhi]-\mathbb{E}_{p_{k}}[\bPhi] - t(1-\alpha)\bw_{k}} \leqslant t\alpha(1+10^{-5}).
\]

\subsubsection*{Non-overlapping group lasso Maxent model}
For this Maxent model, we divide the features into five disjoint groups of features, as described in~Supplement~\cref{supp:wumi_features}. Then, the smallest hyperparameter for which the solutions to the primal and dual problems \cref{eq:maxent_EN} and \cref{eq:maxent_EN_dual} are equal to the prior distribution and zero is
\[
t^{(0)} = \max_{g \in \{1,\dots,5\}} \left\{\normg{\evd[\bPhi_{g}] - \evp[\bPhi_{g}]}/\sqrt{m_{g}}\right\},
\]
where $m_{g}$ is the number of features in the $g^{\text{th}}$ group. This follows from the optimality condition
\begin{equation}\label{eq:opt_cond_group}
\evd[\bPhi] - \mathbb{E}_{p^{s}}[\bPhi] \in \bigcup_{g=1}^{5} \left\{t\sqrt{m_{g}}\partial\normg{\cdot}\right\}\bw_{g}^{s}
\end{equation}
and computing the smallest parameter $t$ for which \cref{eq:opt_cond_group} is satisfied at $(p^{s},\bw^{s}) = (p_{\text{prior}},\boldsymbol{0})$.

For this model, we use the NPDHG algorithm \cref{eq:NPDHG_g-scheme_nonsmooth_explicit}. Starting from $l = 1$, we compute the pair of solutions $(p^{(l)},\bw^{(l)})$ to the primal and dual problems \cref{eq:maxent_group} and \cref{eq:maxent_group_dual} at hyperparameter $t = t^{(l)}$ using the previously computed pair of solutions $(p^{(l-1)},\bw^{(l-1)})$ by setting the initial vectors $\bz_{0} = \bw_{0} = \bw^{(l-1)}$. For the stepsize parameters, we set $\theta_{0} = 0$, $\tau_{0} = 2$ and $\sigma_{0} = 1/2\normop{\matr{\norm{A}}}^{2}$. Finally, we compute the update $\bw_{k+1}$ in \cref{eq:NPDHG_g-scheme_nonsmooth_explicit} using the following proximal operator formula: for every group $g \in \{1.\dots,5\}$, $\lambda > 0$ and $\hat{\bw} \in \Rm$,
\begin{equation*}
    \argmin_{\bw \in \Rm} \left\{\lambda  \sqrt{m_{g}}\normg{\bw} + \frac{1}{2}\normg{\bw-\hat{\bw}}^{2}\right\} =  \max{\left(0,1-\lambda\sqrt{m_{g}}/\normtwo{\hat{\bw}}\right)}\hat{\bw}.
\end{equation*}
Finally, we let the NPDHG algorithms run for at least 40 iterations before checking for convergence. We stop the NPDHG algorithms when the optimality condition \cref{eq:opt_cond_group} is satisfied within some tolerance $10^{-5}$:
\[
\max_{g \in \{1,\dots,5\}} \normg{\evd[\bPhi] - \mathbb{E}_{p_{k}}[\bPhi]} \leqslant t(1+10^{-5}).
\]

\subsubsection*{$\ell_{\infty}$-regularized Maxent model}
For this Maxent model, the smallest hyperparameter for which the solutions to the primal and dual problems \cref{eq:maxent_EN} and \cref{eq:maxent_EN_dual} are equal to the prior distribution and zero is
\[
t^{(0)} = \normone{\evd[\bPhi]-\evprior[\bPhi]}.
\]
This follows from the optimality condition
\begin{equation}\label{eq:opt_cond_linf}
\evd[\bPhi] - \mathbb{E}_{p^{s}}[\bPhi] \in t\partial\norminf{\cdot}(\bw^{s})
\end{equation}
and computing the smallest parameter $t$ for which \cref{eq:opt_cond_linf} is satisfied at $(p^{s},\bw^{s}) = (p_{\text{prior}},\boldsymbol{0})$.

For this model, we use the NPDHG algorithm \cref{eq:NPDHG_g-scheme_nonsmooth_explicit}. Starting from $l = 1$, we compute the pair of solutions $(p^{(l)},\bw^{(l)})$ to the primal and dual problems \cref{eq:maxent_linf} and \cref{eq:maxent_linf_dual} at hyperparameter $t = t^{(l)}$ using the previously computed pair of solutions $(p^{(l-1)},\bw^{(l-1)})$ by setting the initial vectors $\bz_{0} = \bw_{0} = \bw^{(l-1)}$. For the stepsize parameters, we set $\theta_{0} = 0$, $\tau_{0} = 2$ and $\sigma_{0} = 1/2\normop{\matr{\norm{A}}}^{2}$. Finally, we compute the update $\bw_{k+1}$ in \cref{eq:NPDHG_g-scheme_nonsmooth_explicit} using Moreau's decomposition~\cite[Theorem 3.2.5]{hiriart2013convexI}: for every $\lambda > 0$ and $\hat{\bw} \in \Rm$,
\begin{equation*}
    \argmin_{\bw \in \Rm} \left\{\lambda  \norminf{\bw} + \frac{1}{2}\normtwo{\bw-\hat{\bw}}^{2}\right\} = \hat{\bw} - \argmin_{\normone{\bw} \leqslant \lambda} \frac{1}{2}\normtwo{\bw-\hat{\bw}}^{2}.
\end{equation*}
The second term on the right amounts to projecting $\hat{\bw}$ on the $\ell_{1}$ ball of radius $\lambda$. There are fast algorithms for doing do; we use Algorithm 1 described in~\cite{condat2016fast}. Finally, we let the NPDHG algorithms run for at least 40 iterations before checking for convergence. We stop the NPDHG algorithms when the optimality condition \cref{eq:opt_cond_linf} is satisfied within some tolerance $10^{-5}$:
\[
\normone{\evd[\bPhi]-\mathbb{E}_{p_{k}}[\bPhi]} \leqslant t(1+10^{-5}).
\]

\subsection{Comparison of timings}\label{subsec:numerics_comparison}
In this section, we compare the run times of our NPDHG algorithms with two state-of-the-art optimization algorithms for solving nonsmooth Maxent models: the forward-backward splitting algorithm (specifically, Algorithm 5 in~\cite{chambolle2016introduction}; see also~\cite{beck2009fast,daubechies2004iterative}) and the STRUCTMAXENT2 coordinate descent algorithm from~\cite{cortes2015maxent,mohri2018foundations}. All numerical experiments were performed on a single core Intel(R) Core(TM) i7-10750H CPU @ 2.60 GHz.

We initially chose the GLMNET implementation~\cite{fithian2013finite,phillips2017opening} in MATLAB over STRUCTMAXENT2 for the numerical comparisons, but we found that GLMNET produced unreliable numerical solutions when compared to both the NPDHG and forward-backward splitting algorithms. We also tried using GLMNET's implementation in the R language, but to no avail. We think this problem arises because, as discussed in \cref{sec:sota_methods}, the GLMNET algorithm approximates a Maxent model as a logistic regression model and then invokes a coordinate descent method tailored to logistic regression to approximate the solution to the Maxent model. Our observations suggest this approach does not work well for our data set. In contrast, we found that the STRUCTMAXENT2 algorithm produced correct numerical results.

For the forward-backward splitting algorithm, the stepsize parameters were set to $1/\normtwo{\matr{A}}$ and $1$ (corresponding to $\tau$ and $t_{0}$ in Algorithm 5 of~\cite{chambolle2016introduction}) and for computing the pair of solutions $(p^{(l)},\bw^{(l)})$ at hyperparameter $t^{(l)}$, the initial iterate was set to $\bw^{(l-1)}$. In addition, for the elastic net Maxent models, the acceleration quantity $q = (1-\alpha)t^{(l)}/(\normtwo{\matr{A}} + (1-\alpha)t^{(l)})$ was employed. We used the stopping criteria of the NPDHG algorithms for the forward-backward splitting algorithms. For the coordinate descent algorithm, we modified the STRUCTMAXENT2 algorithm from~\cite[Page 305-306]{mohri2018foundations} to make it applicable to the elastic net penalty. For computing the pair of solutions $(p^{(l)},\bw^{(l)})$ at hyperparameter $t^{(l)}$, the initial iterate was set to $\bw^{(l-1)}$. We did not use the STRUCTMAXENT2 algorithm for the non-overlapping group lasso or $\ell_{\infty}$-regularized Maxent models, as it was not designed for these Maxent models. We used the stopping criteria of the NPDHG algorithms for the STRUCTMAXENT2 algorithm.

\cref{tab:numres1} shows the average timings for computing the entire regularization path of the WUMI wildfire data set using the Coordinate descent, the forward-backward splitting and NPDHG algorithms. All timings were averaged over five runs, and for the Forward-backward splitting and NPDHG algorithms, they include the time required to compute all the stepsize parameters. All algorithms were implemented in MATLAB. 
\begin{table}[!h]
    \centering
    \begin{tabular}{|c|c|c|c|}
    \hline
            & STRUCTMAXENT2 & Forward-backward splitting & NPDHG\\
            \hline
            Elastic net ($\alpha = 0.95$) & 5562.19 & 4208.01 & \textbf{365.55} \\\hline

            Elastic net ($\alpha = 0.40$) & 1018.73 & 1407.22 & \textbf{113.53} \\\hline
            
            Non-overlapping group lasso & N/A & 3036.38 & \textbf{278.14} \\\hline
            
            $\ell_{\infty}$-regularization & N/A & 2534.65 & \textbf{289.98} \\\hline
    \end{tabular}
    \caption{Timings results (in seconds) for fitting the Maxent models described in \cref{subsec:fitting_setup}. All timings are averaged over five runs.}
    \label{tab:numres1}
\end{table}

The NPDHG algorithm outperformed both the forward-backward splitting and coordinate descent algorithms by at least one order of magnitude. In particular, we observed that the NPDHG algorithm required far fewer iterations to achieve convergence compared to both the forward-backward splitting and STRUCTMAXENT2 algorithms. This difference is because the stepsize parameters for the NPDHG algorithm were much larger compared to either the forward-backward splitting and STRUCTMAXENT2 algorithm. Indeed, the stepsize parameters for the NPDHG and forward-backward splitting algorithms are inversely proportional to the norms $\normop{A}$ and $\normtwo{A}$, and for the wildfire data set these were $\normop{\matr{A}} \approx 3.30$ and $\normtwo{A} \approx 854.08$. Thus larger stepsize parameters were permitted thanks to the Kullback--Leibler divergence term in the NPDHG algorithm, enabling a major speedup gain.

\begin{figure}[!h]
\centering
\includegraphics[width= 0.464\textwidth]{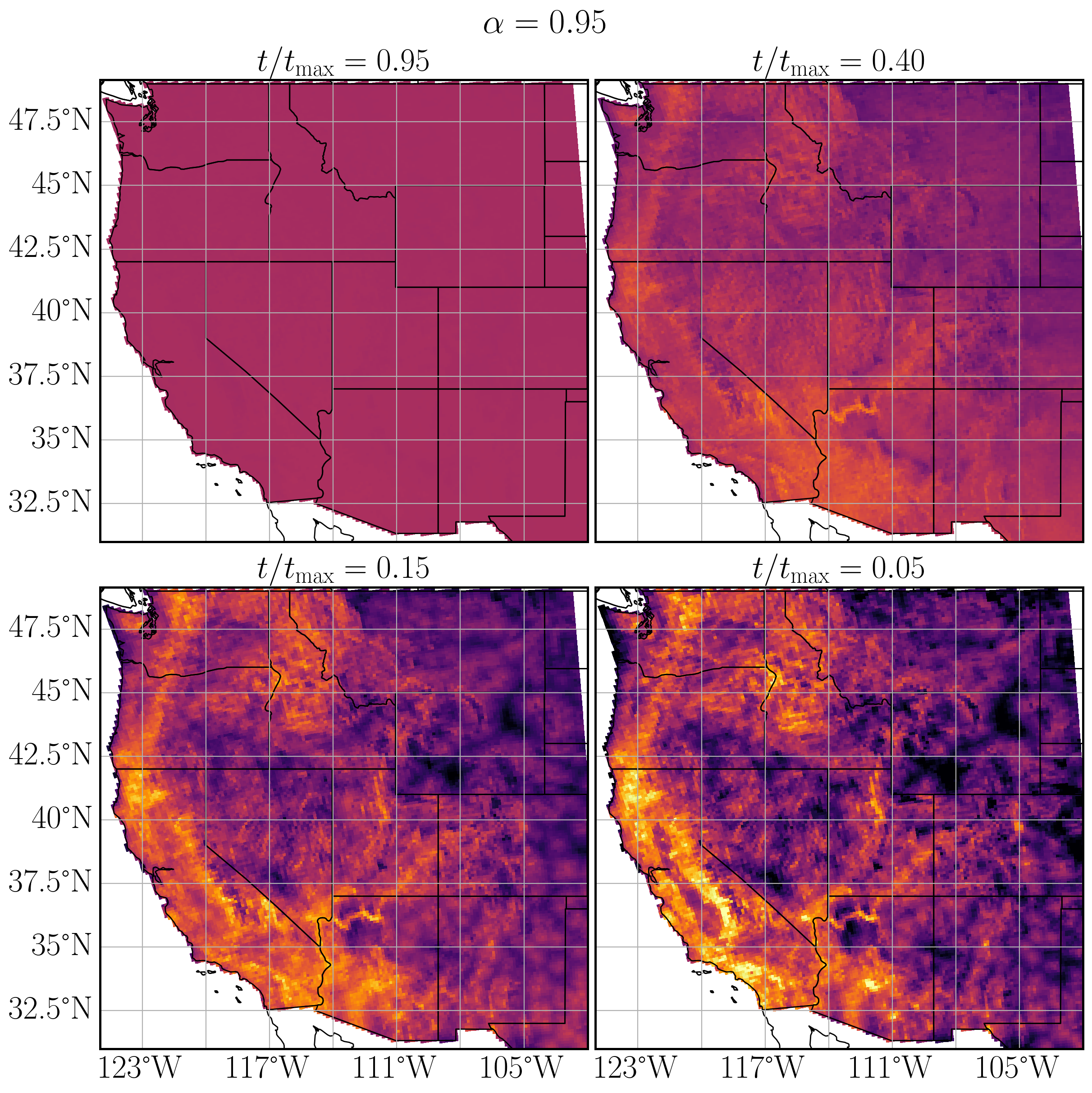}
\includegraphics[width= 0.52\textwidth]{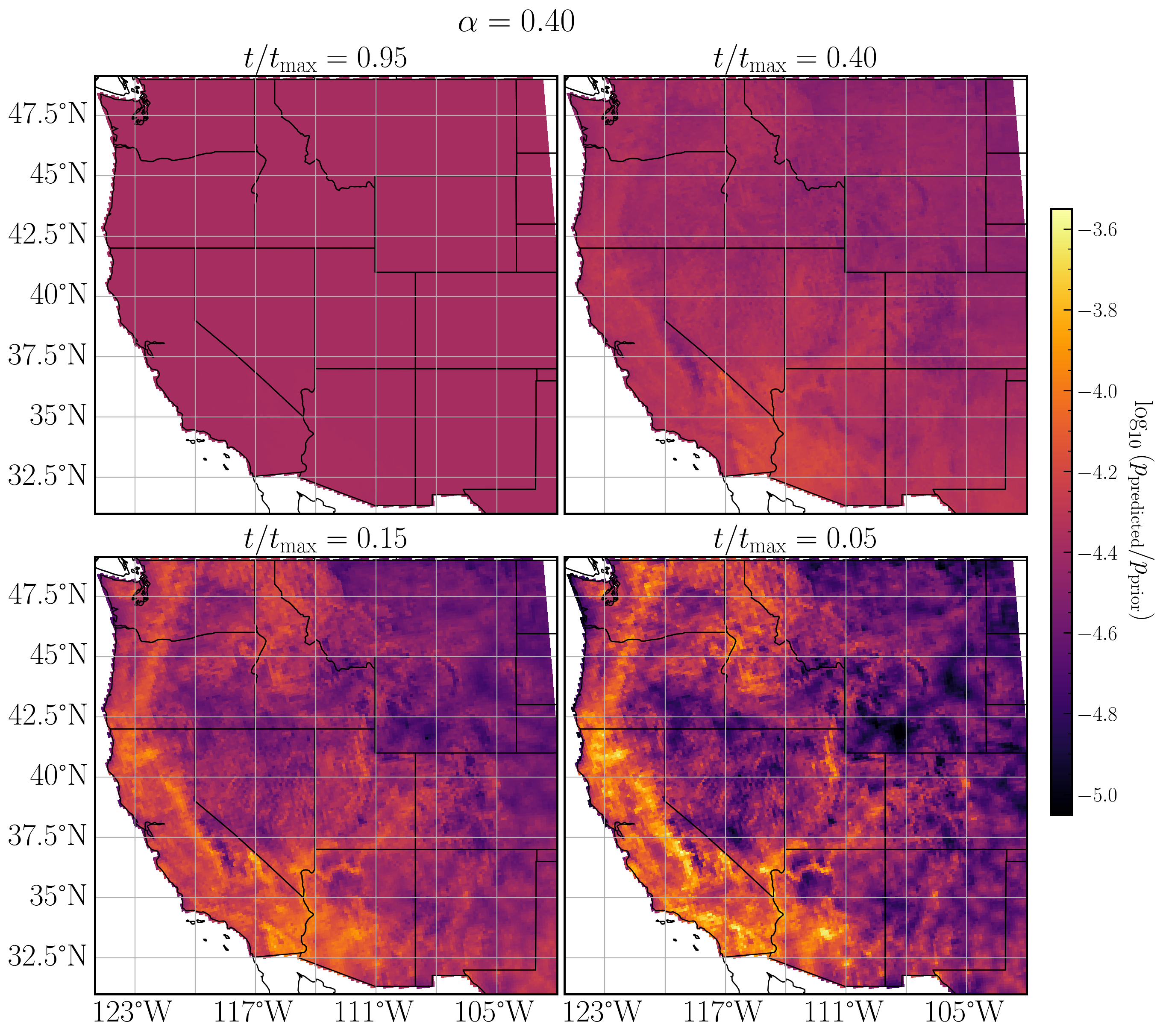}
\includegraphics[width= 0.464\textwidth]{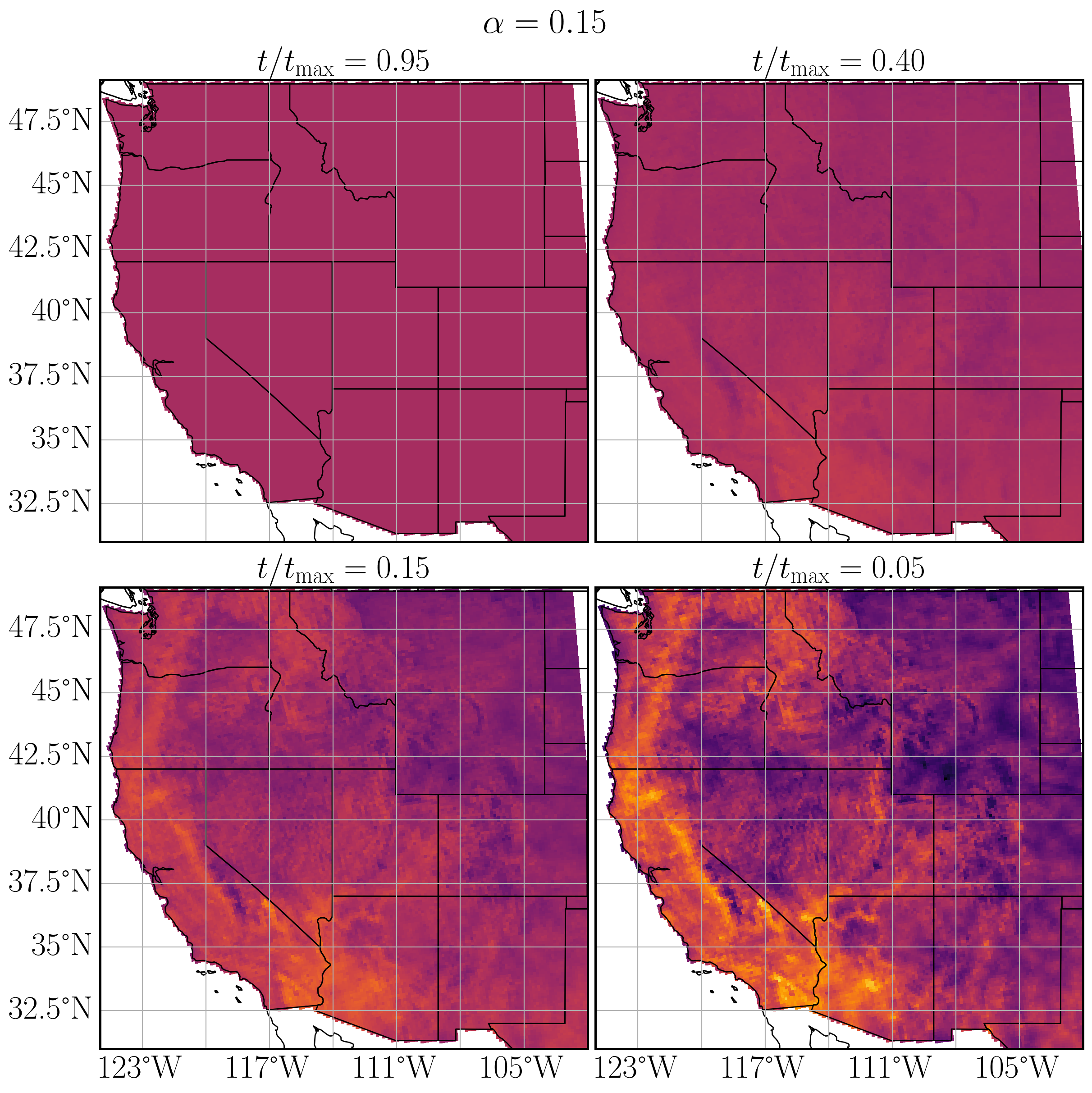}
\includegraphics[width= 0.52\textwidth]{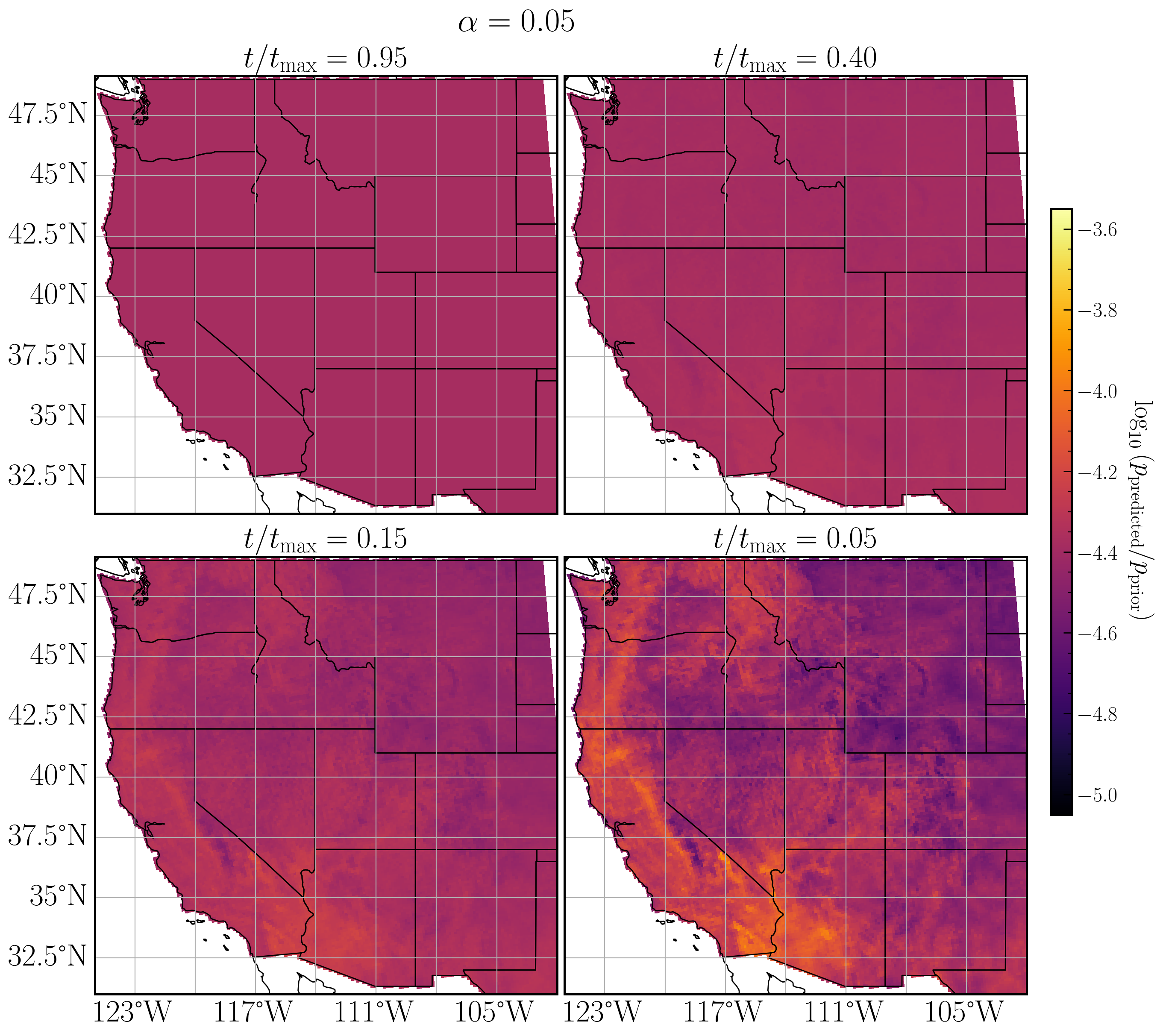}
\caption{Spatial probability plot for different hyperparameter values with elastic net penalty parameter $\alpha = \{0.95, 0.40, 0.15, 0.05 \}$.}
\label{fig:hypvar_elastic_net}
\end{figure}

\subsection{Analysis of the Maxent regularization paths and estimated fire probabilities}\label{subsec:numerics_analysis}

\begin{figure}[!h]
\centering
\includegraphics[width= 0.465\textwidth]{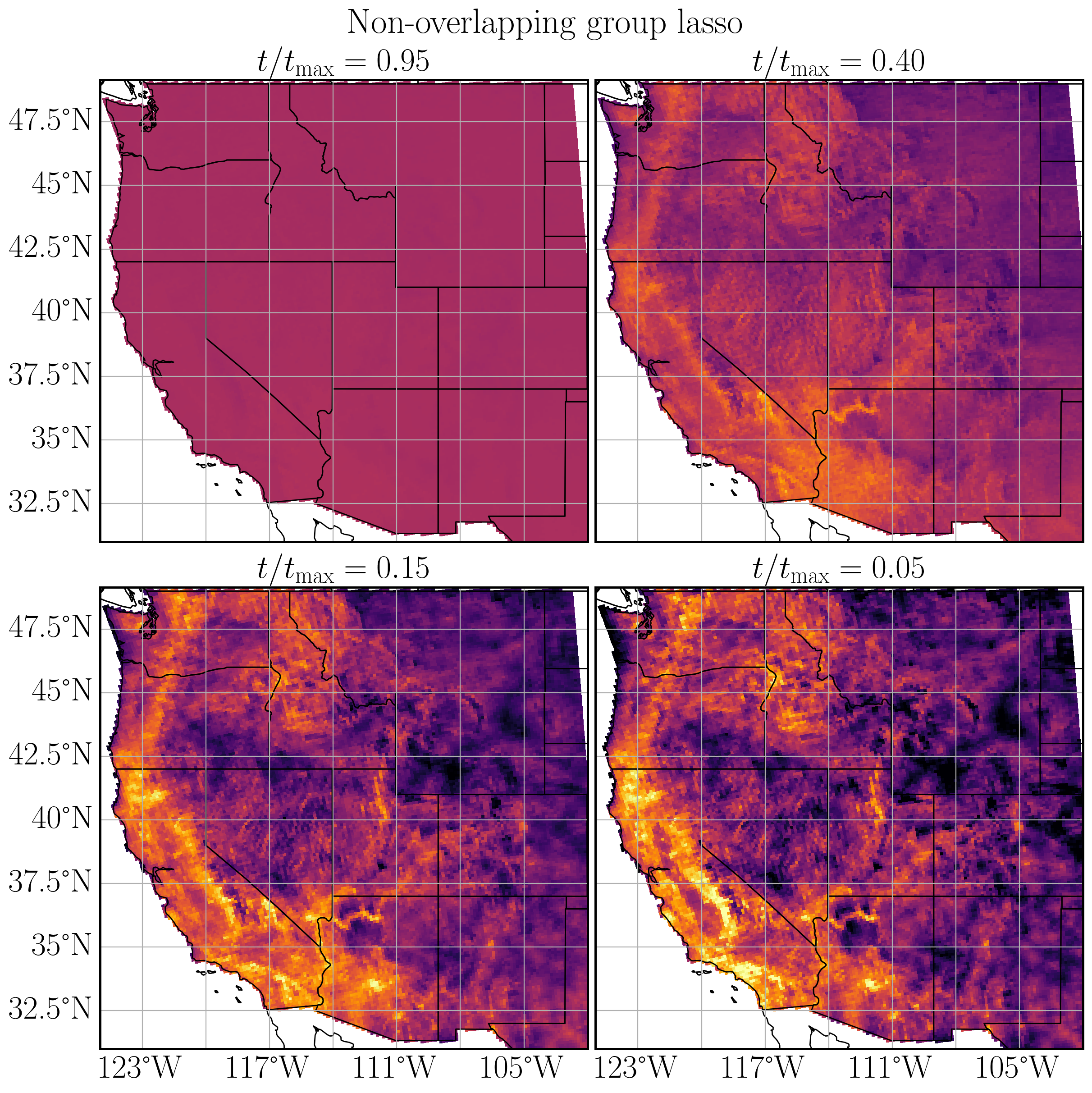}
\includegraphics[width= 0.52\textwidth]{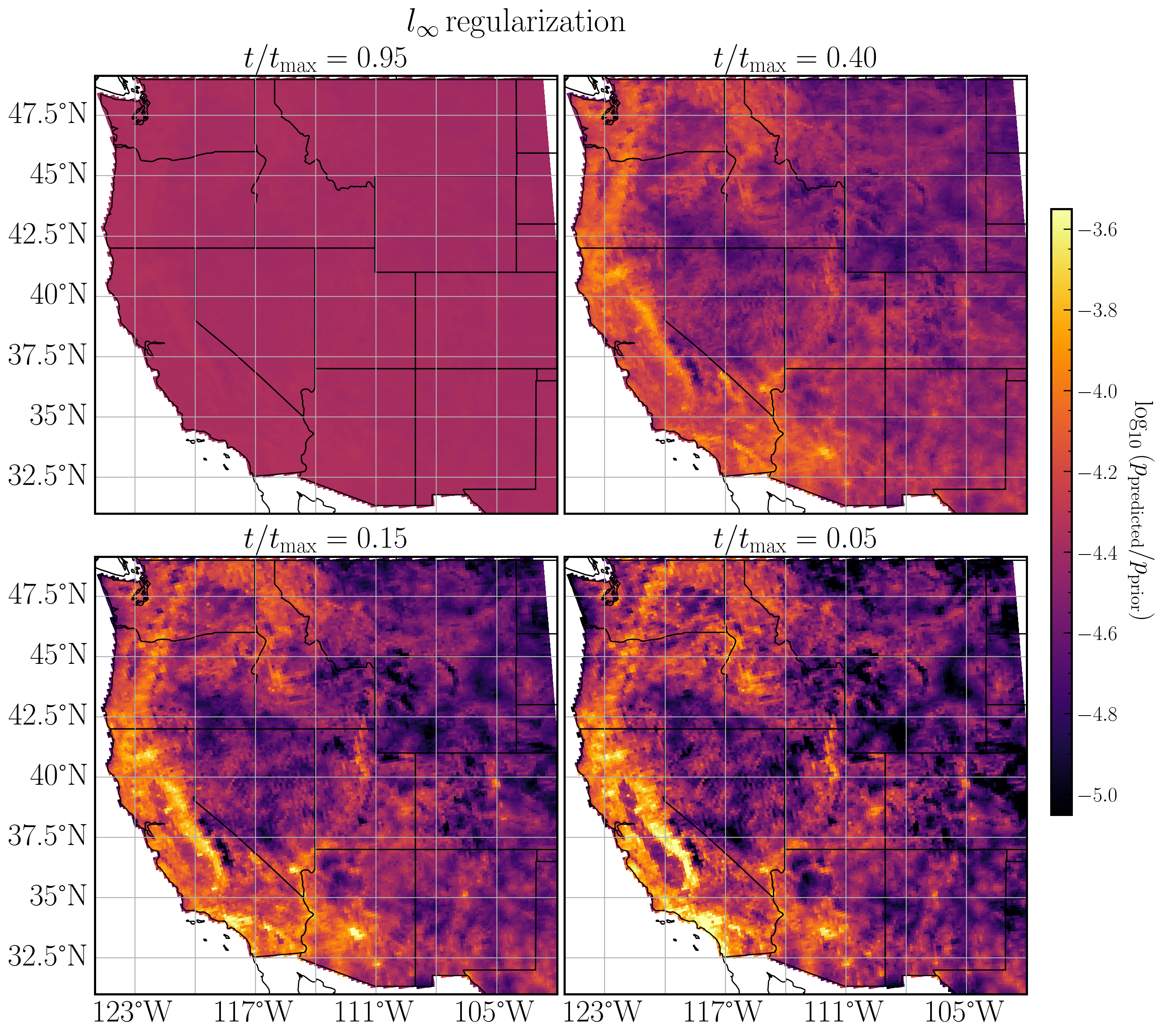}
\caption{Same as \cref{fig:hypvar_elastic_net} but for (left) the non-overlapping group lasso with $\alpha = 1$, and (right) the $l_\infty$ Maxent models respectively.}
\label{fig:hypvar_nogl_linf}
\end{figure}

%As a final validation step for the NPDHG algorithm, we visualize the normalized mean fire probability across the western US in \cref{fig:hypvar_elastic_net} and \cref{fig:hypvar_nogl_linf}. Here, we use the fitted Maxent models to compute the fire probability in each grid cell for all calendar months between 1984 and 2020. In each case, we have chosen the prior distribution as a benchmark for our spatial plots of fire probability. 

As a final validation step for the NPDHG algorithm, we use the fitted Maxent models to compute the normalized mean fire probability in each grid cell for all calendar months between 1984 and 2020. The probabilities are visualized in  \cref{fig:hypvar_elastic_net} and \cref{fig:hypvar_nogl_linf}. In each case, we have chosen the prior distribution as a benchmark for our spatial plots of fire probability. 

The spatial fire probability for $\alpha = \{0.95, 0.40, 0.15, 0.05 \}$ are shown in \cref{fig:hypvar_elastic_net}. The range of $\alpha$ values roughly corresponds to varying the regularization from a purely $l_1$ norm ($\alpha = 1 $) to a purely $l_2$ norm ($\alpha = 0$). For each value of $\alpha$, we also consider the evolution of the spatial fire probability as we vary the hyperparameter $t$ along the regularization path, or equivalently include additional features while fitting the Maxent model to wildfire data. Broadly, we observe that for a fixed value of $t/t_{\rm max}$, the ratio of predicted to prior fire probability decreases in sharpness as $\alpha$ decreases. On the other hand, for a fixed $\alpha$, decreasing $t/t_{\rm max}$ values enables the model to make sharper distinctions between grid cells with high and low fire probability as evidenced by the sharper contrast between the prior and predicted fire probabilities. We also note a similar pattern in \cref{fig:hypvar_nogl_linf} for the Maxent models with non-overlapping group lasso (corresponding roughly to the elastic net case with $\alpha= 1$) and the $l_\infty$-Maxent models, all which converge to the empirical distribution quicker than any of the elastic net cases. 

\begin{figure}[!ht]
\centering
\includegraphics[width= 0.84\textwidth]{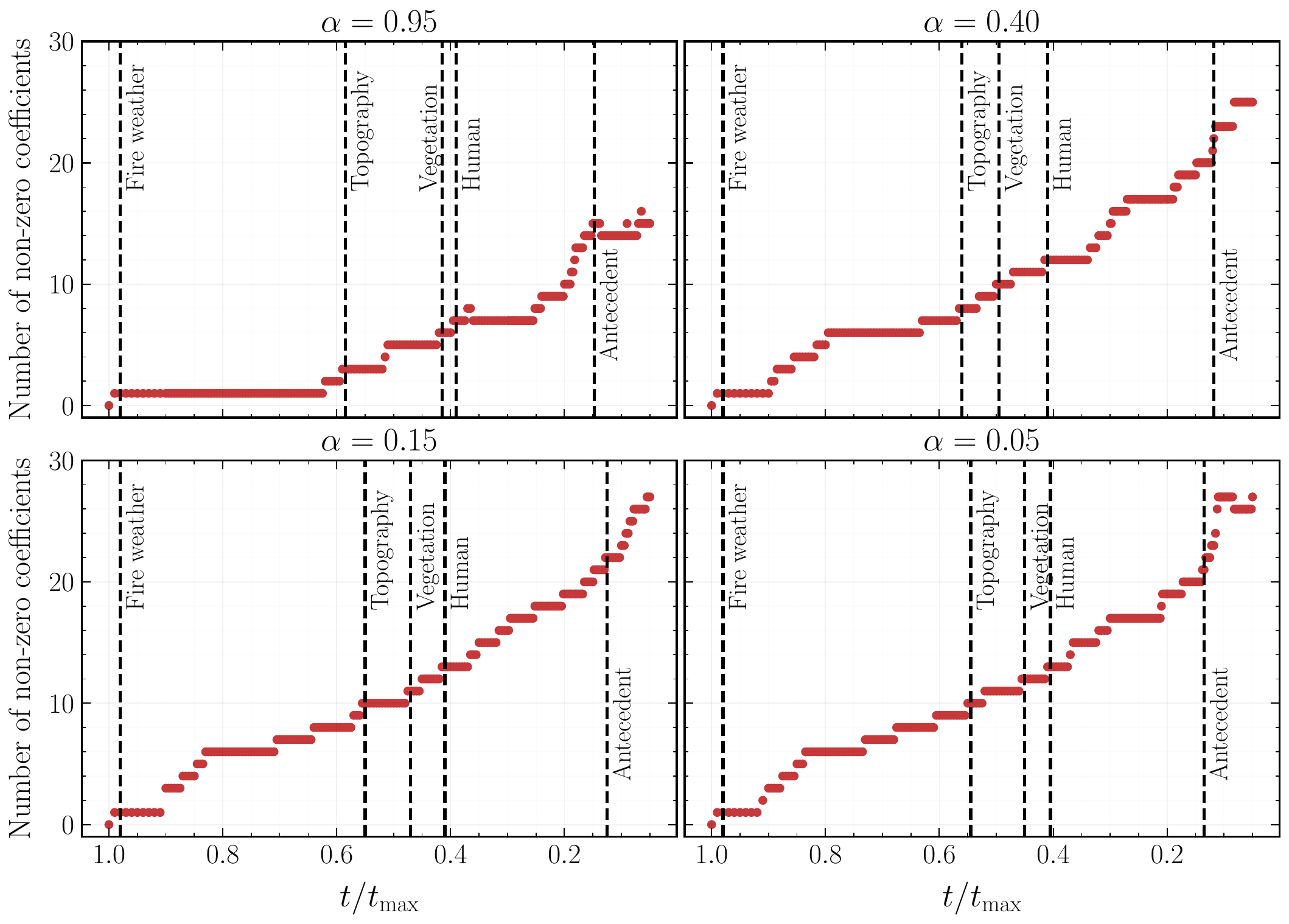}
\caption{Number of non-zero coefficients along the regularization path plots for elastic net penalty parameter $\alpha = \{0.95, 0.40, 0.15, 0.05 \}$. The dashed vertical lines highlight the $t/t_{\rm max}$ value at which the first feature of the group indicated by inset text is selected.}
\label{fig:regpath_en}
\end{figure}

\begin{table}[!ht]
    \centering
    \begin{tabular}{c|c || c|c|}
        \hline
        Feature group & $\alpha=0.95$  & \multicolumn{2}{c|}{$\alpha=0.05$}  \\ \hline
        \multirow{8}{*}{Fire weather} & Tmax & Tmax & VPD \\[0.25em]
        & Prec & Prec & Tmin  \\[0.25em]
        & Wind & Wind & ${\rm VPD}_{\rm max3}$  \\[0.25em]
        & FM1000 & FM1000 & ${\rm VPD}_{\rm max7}$  \\[0.25em]
        & ${\rm Tmin}_{\rm max3}$ & ${\rm Tmin}_{\rm max3}$ & ${\rm Tmax}_{\rm max3}$  \\[0.25em]
        & Lightning & Lightning & ${\rm Tmax}_{\rm max7}$  \\[0.25em]
        & & ${\rm Tmin}_{\rm max7}$ & ${\rm SWE}_{\rm max}$ \\[0.25em]
        \hline
        \multirow{2}{*}{Topography} & Slope & Slope & Southness  \\[0.25em]
        & Southness & & \\[0.25em] \hline
        \multirow{2}{*}{Vegetation} & Grassland & Biomass & Forest \\[0.25em] 
        & & Shrub & \\ \hline
        \multirow{4}{*}{Human} & Camp\textunderscore dist & Camp\textunderscore dist & Road\textunderscore dist \\[0.25em] 
         & Camp\textunderscore num & Camp\textunderscore num & \\[0.25em] 
         & Urban & Urban &  \\[0.25em] 
         & Popdensity & Popdensity & \\[0.25em] \hline
        \multirow{1}{*}{Antecedent} & ${\rm AvgPrec}_{\rm 2mo}$ & ${\rm AvgPrec}_{\rm 2mo}$ & ${\rm AntPrec}_{\rm lag2}$  \\[0.25em] & ${\rm AvgVPD}_{\rm 3mo}$ & ${\rm AntPrec}_{\rm lag1}$ & \\[0.25em] \hline
    \end{tabular}
    \caption{List of non-zero features at the end of the regularization path for two elastic net penalty parameters, organized by different feature groups. See \cref{supp:wumi_features} for additional description of the selected features.}
    \label{tab:feature_selection}
\end{table}

In \cref{fig:regpath_en}, we show the cumulative number of non-zero coefficients at fixed intervals along the regularization path for different $\alpha$ values. The plot helps in visualizing the $t/t_{\rm max}$ values at which new features are introduced in the elastic net Maxent model, with the dashed vertical lines indicating the first time a feature from a new group is selected. Across all $\alpha$ values we find that features appear in the same order, with fire weather features being selected first, followed by topography, vegetation, human, and antecedent features. We tabulate the final set of non-zero features at the end of the regularization path for $\alpha= \{ 0.95, 0.05 \}$ across various groups in Table~\ref{tab:feature_selection}. These selected features are in good agreement with physical models of wildfire occurrence~\cite{Parisien_2019, Mezuman2020} as well as previous statistical analyses of wildfire drivers~\cite{Parisien2009, chen2021,yu_et_al_2021, Buch2023}.

% Sparsity plots for alpha = 0.95 and 0.05
% Table for alpha = 0.95: List the nonzero features. Put the parameters in order that they appear.
\section{Discussion}\label{sec:conclusion}
In this paper, we have introduced novel first-order NPDHG algorithms that overcome the shortcomings of state-of-the-art algorithms for training large-scale, non-smooth Maxent models. The crucial ingredient common to our algorithms is the Kullback--Leibler divergence. Using it over the classical proximal operator makes it possible to train large-scale and non-smooth Maxent models much more efficiently than the state of the arts. In particular, all stepsize parameters and iterations in our algorithms can be calculated on the order $O(mn)$ operations, improving on the complexity bound of $O(\min(m^2n,mn^2))$ operations for computing the optimal stepsize parameters of classical first-order optimization methods, such as the linear PDHG or forward-backward splitting methods. As a consequence, for a given tolerance level $\epsilon > 0$, our algorithms provably compute solutions using on the order of $O(mn/\sqrt{\epsilon})$ or $O(mn/\log(1/\epsilon))$ operations, the order depending on the smoothness of the Maxent model and which are optimal with respect to the Nesterov class of optimal first-order methods~\cite{nesterov2004lectures}. Moreover, the computational bottleneck consists of matrix-vector multiplications, which can be trivially parallelized, and so our algorithms exhibit scalable parallelism. 

Finally, we have shown that the strong convexity of the Kullback--Leibler divergence with respect to the $\ell_{1}$ norm allows for significantly larger stepsize parameters, thereby speeding up the convergence rate of our algorithms. This was, in particular, observed in \cref{sec:maxent_numerics}, when we applied our algorithms to fit the WUMI wildfire data set~\cite{Juang2022} on several non-smooth Maxent models to estimate probabilities of fire occurrences as a function of ecological features. Our algorithms outperformed the state-of-the-art forward-backward splitting and coordinate descent STRUCTMAXENT2 algorithms by at least one order of magnitude. They also yielded results that are in good agreement with physical models of wildfire occurrence~\cite{Parisien_2019, Mezuman2020} as well as previous statistical analyses of wildfire drivers~\cite{Parisien2009, chen2021,yu_et_al_2021, Buch2023}. Future work will also explore the scalability of our algorithms for modeling daily scale wildfire probability~\cite{keeping_modelling_2024}.

We expect our algorithms to provide efficient methods for solving non-smooth Maxent models that arise in large-scale machine learning applications beyond the wildfire application explored in this paper. It would be interesting and impactful to extend our algorithms to continuous regularized Maxent models. Interesting, because the continuous version of the Maxent problem is posed over a non-reflexive space, which makes this problem more technically challenging. Impactful, because such an algorithm would enable a much broader class of continuous probability distributions to be used in Maxent modelling.

\section*{Conflict of interest}
The authors declare that they have no conflict of interest.

\section*{Funding Information}
GPL and JD are supported by ONR N00014-22-1-2667. JD is also supported by DOE-MMICS SEA-CROGS DE-SC0023191. JB is supported by the Zegar Family Foundation and the National Science Foundation LEAP Science and Technology Center Award \#2019625.

\clearpage
\appendix
\section{Derivation of update~\eqref{eq:update_pkplusone}}\label{supp:update-p}
We divide the derivation into two parts, first deriving an expression for the update $p_{k+1}$ in terms of the previous iterate $p_{k}$ and then deriving the explicit update for $p_{k+1}$ as in the third line of~\eqref{eq:NPDHG_g-scheme_nonsmooth}.

\bigbreak
\noindent
\textbf{Part 1.} First, note the minimization problem
\[
\min_{p \in \Delta_{n}} \left\{\tau_{k}\kldiv{p}{p_{\text{prior}}} - \tau_{k}\left\langle\bw_{k} + \theta_{k}(\bw_{k}-\bw_{k-1}),\evp[\bPhi]\right\rangle + \kldiv{p}{p_{k}} \right\}
\]
has a unique global minimum, denoted by $p_{k+1}$, in the relative interior of the unit simplex $\Delta_{n}$ for every $k \in \mathbb{N}$. See~\cite[Proposition 4.1]{langlois2021accelerated} for details and note that Proposition 4.1 applies because the Kullback--Leibler divergence is strongly convex with respect to the $\ell_{1}$ norm in its first argument. 

Having established that the minimum $p_{k+1}$ exists and is unique, we introduce the Karush-Kuhn-Tucker multipliers $\xi \in \R$ and $\boldsymbol{\mu} \in \Rn$ to write the minimization problem for $p_{k+1}$ in terms of an unconstrained optimization problem: 
\[
\min_{\substack{p \in \Rn \\ \xi \in \R \\ \mu \in \Rn}} \left\{\tau_{k}\kldiv{p}{p_{\text{prior}}} - \tau_{k}\left \langle\bw_{k} + \theta_{k}(\bw_{k}-\bw_{k-1}),\evp[\bPhi]\right\rangle + \kldiv{p}{p_{k}} + \xi\left(1 - \sum_{j=1}^{n}p(j)\right) - \langle \boldsymbol{\mu},p \rangle\right\}.
\]
We can invoke the Karush--Kuhn--Tucker Theorem and use the linearity constraint qualifications to find that $p_{k+1}$ satisfies the first-order optimality condition
\begin{equation}\label{eq:appA_technical}
\begin{alignedat}{1}
\tau_{k}\log(p_{k+1}(j)/p_{\text{prior}}(j)) + \tau_{k} &-\tau_{k}\langle\bw_{k} + \theta_{k}(\bw_{k}-\bw_{k-1}),\bPhi(j)\rangle \\
&+ \log(p_{k+1}(j)/p_{k}(j)) + 1 -\xi -[\boldsymbol{\mu}]_{j} = 0
\end{alignedat}
\end{equation}
for each $j \in \{1,\dots,n\}$~\cite{boyd2004convex}. Moreover, the complementary slackness condition $\langle \boldsymbol{\mu},p_{k+1}\rangle = 0$ with the fact that $p_{k+1} > 0$ for every $j \in \{1,\dots,m\}$ implies $\boldsymbol{\mu} = \boldsymbol{0}$.

Next, we use the first-order optimality condition to compute $p_{k+1}(j)$ explicitly. Rearranging~\eqref{eq:appA_technical} in terms of $\log(p_{k+1}(j))$, using $\boldsymbol{\mu} = 0$, and taking the exponential yield
\[
p_{k+1}(j) = \left(e^{(\xi-1)/(1+\tau_{k}) - 1}\right)\left((p_{\text{prior}}(j))^{\tau_{k}}p_{k}(j)e^{\tau_{k}\langle\bw_{k} + \theta_{k}(\bw_{k}-\bw_{k-1}),\bPhi(j)\rangle}\right)^{1/(1+\tau_{k})}
\]
Since $\sum_{j=1}^{n} p_{k+1}(j) = 1$, we find that $\xi$ satisfies the relation
\[
e^{\tau_{k}\xi/(1+\tau_{k}) - 1} = \frac{1}{\sum_{j=1}^{n}\left((p_{\text{prior}}(j))^{\tau_{k}}p_{k}(j)e^{\tau_{k}\langle\bw_{k} + \theta_{k}(\bw_{k}-\bw_{k-1}),\bPhi(j)\rangle}\right)^{1/(1+\tau_{k})}}.
\]
Hence
\[
p_{k+1}(j) = \frac{\left((p_{\text{prior}}(j))^{\tau_{k}}p_{k}(j)e^{\tau_{k}\langle\bw_{k} + \theta_{k}(\bw_{k}-\bw_{k-1}),\bPhi(j)\rangle}\right)^{1/(1+\tau_{k})}}{\sum_{j=1}^{n}\left((p_{\text{prior}})^{\tau_{k}}(j)p_{k}(j)e^{\tau_{k}\langle\bw_{k} + \theta_{k}(\bw_{k}-\bw_{k-1}),\bPhi(j)\rangle}\right)^{1/(1+\tau_{k})}}
\]
for each $j \in \{1,\dots,n\}$. This yields an explicit update for $p_{k+1}$ in terms of the previous iterate $p_{k}$.

\bigbreak
\noindent
\textbf{Part 2.} We now induct on $k$ to prove that the explicit expression for the update $p_{k+1}$ in the third line of~\eqref{eq:NPDHG_g-scheme_nonsmooth} is correct. For $k = 0$, we have % State the update here?
\[
p_{0}(j) = \frac{p_{\text{prior}}(j)e^{\langle\bz_{0},\bPhi(j)\rangle}}{\sum_{j=1}^{n}p_{\text{prior}}(j)e^{\langle\bz_{0},\bPhi(j)\rangle}}
\]
and so
\[
\begin{alignedat}{1}
&\left(\left(p_{\text{prior}}(j)\right)^{\tau_{0}}p_{0}(j)e^{\tau_{0}\langle \bw_{0} + \theta_{0}(\bw_{0}-\bw_{-1}),\bPhi(j)\rangle}\right)^{1/(1+\tau_{0})} \\
&\qquad\qquad\qquad\qquad\qquad\qquad = \frac{p_{\text{prior}}(j)e^{\langle(\bz_{0} + \tau_{0} (\bw_{0} + \theta_{0}(\bw_{0}-\bw_{-1})))/(1+\tau_{0}),\bPhi(j)\rangle}}{\left(\sum_{j=1}^{n}p_{\text{prior}}(j)e^{\langle\bz_{0},\bPhi(j)\rangle}\right)^{1/(1+\tau_{0})}}.
\end{alignedat}
\]
Writing
\[
\bz_{1} = (\bz_{0} + \tau_{0} (\bw_{0} + \theta_{0}(\bw_{0}-\bw_{-1})))/(1+\tau_{0}),
\]
the term inside the exponential on the numerator simplifies to $\langle\bz_{1},\bPhi(j)\rangle$. Hence
\[
\left(\left(p_{\text{prior}}(j)\right)^{\tau_{0}}p_{0}(j)e^{\tau_{0}\langle \bw_{0} + \theta_{0}(\bw_{0}-\bw_{-1}),\bPhi(j)\rangle}\right)^{1/(1+\tau_{0})} = \frac{p_{\text{prior}}(j)e^{\langle\bz_{1},\bPhi(j)\rangle}}{\left(\sum_{j=1}^{n}p_{\text{prior}}(j)e^{\langle\bz_{0},\bPhi(j)\rangle}\right)^{1/(1+\tau_{0})}}.
\]
It follows on substitution that
\[
p_{1}(j) = \frac{p_{\text{prior}}(j)e^{\langle\bz_{1},\bPhi(j)\rangle}}{\sum_{j=1}^{n}p_{\text{prior}}(j)e^{\langle\bz_{1},\bPhi(j)\rangle}}
\]
for every $j \in \{1,\dots,n\}$. 

For arbitrary $k \in \mathbb{N}$, we use the induction hypothesis
\[
p_{k}(j) = \frac{p_{\text{prior}}(j)e^{\langle\bz_{k},\bPhi(j)\rangle}}{\sum_{j=1}^{n}p_{\text{prior}}(j)e^{\langle\bz_{k},\bPhi(j)\rangle}}
\]
to find
\[
\begin{alignedat}{1}
&\left(\left(p_{\text{prior}}(j)\right)^{\tau_{k}}p_{k}(j)e^{\tau_{k}\langle \bw_{k} + \theta_{k}(\bw_{k}-\bw_{k-1}),\bPhi(j)\rangle}\right)^{1/(1+\tau_{k})} \\
&\qquad\qquad\qquad\qquad\qquad\qquad= \frac{p_{\text{prior}}(j)e^{\langle(\bz_{k} + \tau_{k} (\bw_{k} + \theta_{k}(\bw_{k}-\bw_{k-1})))/(1+\tau_{k}),\bPhi(j)\rangle}}{\left(\sum_{j=1}^{n}p_{\text{prior}}(j)e^{\langle\bz_{k},\bPhi(j)\rangle}\right)^{1/(1+\tau_{k})}}.
\end{alignedat}
\]
Writing
\[
\bz_{k+1} = (\bz_{k} + \tau_{k} (\bw_{k} + \theta_{k}(\bw_{k}-\bw_{k-1})))/(1+\tau_{k}),
\]
the term inside the exponential on the numerator simplifies to $\langle\bz_{k+1},\bPhi(j)\rangle$. Hence
\[
\left(\left(p_{\text{prior}}(j)\right)^{\tau_{k}}p_{k}(j)e^{\tau_{0k}\langle \bw_{k} + \theta_{k}(\bw_{k}-\bw_{k-1}),\bPhi(j)\rangle}\right)^{1/(1+\tau_{k})} = \frac{p_{\text{prior}}(j)e^{\langle\bz_{k+1},\bPhi(j)\rangle}}{\left(\sum_{j=1}^{n}p_{\text{prior}}(j)e^{\langle\bz_{k},\bPhi(j)\rangle}\right)^{1/(1+\tau_{k})}}.
\]
It follows on substitution that
\[
p_{k+1}(j) = \frac{p_{\text{prior}}(j)e^{\langle\bz_{k+1},\bPhi(j)\rangle}}{\sum_{j=1}^{n}p_{\text{prior}}(j)e^{\langle\bz_{k+1},\bPhi(j)\rangle}}
\]
for every $j \in \{1,\dots,n\}$, which proves the desired result.

\section{Summary of features extracted from the WUMI wildfire data set}\label{supp:wumi_features}

We provide here a summary of all the features used in our statistical analysis. The features are aggregated to a 12 km spatial resolution during data preprocessing. Considering each feature's $M$ antecedent month average and maximum $X$-day running average components as distinct features, the total number of features considered in our analysis adds up to 35. We organize the features into five groups with each group reflecting a major environmental driver of fire occurrences. These groups are also an important ingredient of the non-overlapping group lasso Maxent model.

%	  Feature group & Identifier & Description & Resolution  & Source   \\ \hline 
%\endfirsthead
%	\hline
\begin{longtable}[c]{c | c | p{13em} | c || c }
	\hline
		Feature group & Identifier & Description & Resolution & Source  \\ \hline 
  \endhead
	     \multirow{ 16}{*} {\makecell{Fire weather}} & VPD & Mean vapor pressure deficit & 5 km & Climgrid~\cite{climgrid_ref}, PRISM~\cite{prism_ref} \\[0.25em]
	     & ${\rm VPD}^{{\rm max}X}$ & Maximum $X$-day running average of VPD; $X \in \{3, 7\}$ & 9 km & UCLA-ERA5~\cite{ucla_era5_wrf}  \\[0.25em]
         & Tmax & Daily maximum temperature & 5 km & Climgrid  \\[0.25em] 
         & ${\rm Tmax}^{{\rm max}X}$ & Maximum $X$-day running average of Tmax & 9 km  & UCLA-ERA5  \\[0.25em]
         & Tmin & Daily minimum temperature & 5 km  & Climgrid  \\[0.25em]
         & ${\rm Tmin}^{{\rm max}X}$ & Maximum $X$-day running average of Tmin  & 9 km  & UCLA-ERA5  \\[0.25em]
         & Prec & Precipitation total & 5 km  & Climgrid \\[0.25em]
         & ${\rm SWE}_{\rm mean}$ & Mean snow water equivalent & 500 m  & NSIDC~\cite{sweref_1} \\[0.25em]
         & ${\rm SWE}_{\rm max}$ & Daily maximum snow water equivalent & 500 m  & NSIDC \\[0.25em]
         & FM1000 & 1000-hour dead fuel moisture & 4 km  & gridMET~\cite{gridmet_2013} \\[0.25em]
         & ${\rm FFWI}^{{\rm max}7}$ & Maximum 7-day running average of Fosberg Fire Weather Index & 9 km  & UCLA-ERA5 \\[0.25em]
         & Wind & Mean wind speed & 9 km  & UCLA-ERA5  \\[0.25em]
         & Lightning & Lightning strike density & 500 m  & NLDN~\cite{lightning_ref1, lightning_ref2}  \\[0.25em]  \hline
        \multirow{ 10}{*} {\makecell{Antecedent}} & ${\rm AntVPD}_{M{\rm mon}}$   & Average VPD in $M$ antecedent months; $M \in \{2, 3\}$ & 5 km &   Climgrid \\[0.25em]
         & ${\rm AntPrec}_{M{\rm mon}}$ & Average precipitation total in $M$ antecedent months; $M \in \{2, 4\}$ & 5 km  & Climgrid  \\[0.25em]
         & ${\rm AntPrec}_{\rm lag1}$ & Mean annual precipitation in lag year 1 & 5 km  & Climgrid  \\[0.25em]
         & ${\rm AntPrec}_{\rm lag2}$ & Mean annual precipitation in lag year 2 & 5 km  & Climgrid  \\[0.25em] 
         & ${\rm AvgSWE}_{3{\rm mon}}$ & Average snow water equivalent in 3 antecedent months & 500 m  & NSIDC \\[0.25em] \hline
		\multirow{4}{*}{Vegetation} & Forest & Fraction of forest landcover & 30 m  & NLCD \\[0.25em]
         & Grassland & Fraction of grassland cover & 30 m  & NLCD  \\[0.25em] 
         & Shrubland & Fraction of shrubland cover & 30 m  & NLCD  \\[0.25em] 
         & Biomass & Aboveground biomass map & 300 m  & Ref.~\cite{spawn_harmonized_2020} \\[0.25em] \hline
        \multirow{ 2}{*}{Human} & Urban & Fraction of land covered by urban areas & 30 m  & NLCD  \\[0.25em] 
        & Camp\textunderscore num & Mean number of camp grounds & 1km  & Open source \\[0.25em]
         & Camp\textunderscore dist & Mean distance from nearest camp ground & 1km  & Open source \\[0.25em] 
         & Road\textunderscore dist & Mean distance from nearest highway & 1km  & Open source  \\[0.25em] 
         & Popdensity & Mean population density & 1km  & SILVIS~\cite{silvis_ref} \\[0.25em] 
         & Housedensity & Mean housing density & 1km  & SILVIS \\[0.25em]  \hline
        \multirow{ 3}{*}{Topography} & Slope & Mean slope & 1m  & USGS \\[0.25em]
         & Southness & Mean south-facing degree of slope & 1m  & USGS \\[0.25em] \hline
\caption{Summary table for all input features organized by group, identifier, description, spatial resolution of raw data, and source of data.} \label{tab:datasummary}
\end{longtable}

\clearpage
\bibliographystyle{siamplain}
\bibliography{proj-bib}
\end{document}